\def\year{2021}\relax
\documentclass[letterpaper]{article} 
\usepackage{aaai21}  
\usepackage{times}  
\usepackage{helvet} 
\usepackage{courier}  
\usepackage[hyphens]{url}  
\usepackage{graphicx} 
\graphicspath{{ms_FinalFig/}}
\usepackage{natbib}  
\usepackage{caption} 
\usepackage{algorithmic}
\usepackage[ruled,vlined]{algorithm2e}
\usepackage{amsmath}
\usepackage{amsthm}
\usepackage{amssymb, bbm}
\usepackage{graphicx}
\usepackage{subcaption}
\usepackage{footmisc}
\usepackage{url}
\usepackage{color}
\usepackage{subfiles}
\usepackage{booktabs}

\newtheorem{theorem}{Theorem}

\theoremstyle{definition}

\theoremstyle{definition}
\newtheorem{definition}{Definition}

\DeclareMathOperator{\E}{\mathbb{E}}
\DeclareMathOperator{\R}{\mathbb{R}}

\title{Variance Penalized On-Policy and Off-Policy Actor-Critic}
\author {
    Arushi Jain,\textsuperscript{\rm 1, \rm 2}
    Gandharv Patil, \textsuperscript{\rm 1, \rm 2}
    Ayush Jain, \textsuperscript{\rm 1, \rm 2}\\
    Khimya Khetarpal,\textsuperscript{\rm 1, \rm 2}
    Doina Precup \textsuperscript{\rm 1, \rm 2, \rm 3}\\
}
\affiliations {
    \textsuperscript{\rm 1} McGill University,
    \textsuperscript{\rm 2} Mila,
    \textsuperscript{\rm 3} Google DeepMind, Montreal. \\
    \{arushi.jain, gandharv.patil, ayush.jain, khimya.khetarpal\}@mail.mcgill.ca, dprecup@cs.mcgill.ca
}

\begin{document}

\maketitle

\begin{abstract}
Reinforcement learning algorithms are typically geared towards optimizing the expected return of an agent. However, in many practical applications, low variance in the return is desired to ensure the reliability of an algorithm. In this paper, we propose on-policy and off-policy actor-critic algorithms that optimize a performance criterion involving both mean and variance in the return. Previous work uses the second moment of return to estimate the variance indirectly. Instead, we use a much simpler recently proposed direct variance estimator which updates the estimates incrementally using temporal difference methods. Using the variance-penalized criterion, we guarantee the convergence of our algorithm to locally optimal policies for finite state action Markov decision processes. We demonstrate the utility of our algorithm in tabular and continuous MuJoCo domains. Our approach not only performs on par with actor-critic and prior variance-penalization baselines in terms of expected return, but also generates trajectories which have lower variance in the return.
\end{abstract}

\section{Introduction}
\label{sec:introduction}

Reinforcement learning (RL) agents learn to solve a task by optimizing the expected accumulated discounted rewards (return) in a conventional setting. However, in risk-sensitive applications like industrial automation, finance, medicine, or robotics, the standard objective of RL may not suffice, because it does not account for the \textit{variability} induced by the return distribution. In this paper, we propose a technique that promotes learning of policies with less variability.

Variability  in sequential decision-making problems can arise from two sources --  the inherent stochasticity in the environment (transition and reward), and imperfect knowledge about the model. The former source of variability is addressed by the \textit{risk-sensitive} Markov decision processes (MDPs) \cite{howard1972risk,heger1994consideration,borkar2001sensitivity,borkar2002q}, whereas the latter is covered by \textit{robust} MDPs \cite{iyengar2005robust,nilim2005robust}. In this work, we address the former source of variability in an RL setup via mean-variance optimization. One could account for mean-variance tradeoffs via maximization of the mean subject to variance constraints (solved using constrained MDPs \cite{altman1999constrained}), maximization of the Sharpe ratio \cite{sharpe1994sharpe}, or incorporation of the variance as a penalty in the objective function \cite{filar1989variance, white1994mathematical}. Here, we use a \textit{variance-penalized method} to solve the optimization problem by adding a penalty term to the objective.

There are two ways to compute the variance in the return $\text{Var}(G)$. The \textit{indirect approach} estimates $\text{Var}(G)$ using the Bellman equation for both the first moment (i.e. value function) and the second moment as $\text{Var}(G) = \E[G^2] - \E[G]^2$ \cite{sobel1982variance}. The \textit{direct approach} forms a Bellman equation for the variance itself, as $\text{Var}(G) = \E[(G - \E[G])^2]$ \cite{sherstan2018directly}, skipping the calculation of the second moment. \citeauthor{sherstan2018directly} \shortcite{sherstan2018directly} empirically established that in the policy evaluation setting, the direct variance estimation approach is better behaved compared to the indirect approach, in several scenarios: (a) when the value estimates are noisy, (b) when eligibility traces are used in the value estimation, and (c) when  the variance in return is estimated from off-policy samples. Due to the above benefits and the simplicity of the direct approach, we build upon the approach proposed by \citeauthor{sherstan2018directly} \shortcite{sherstan2018directly} only for policy evaluation setting, and, develop actor-critic algorithms for both on- and off-policy settings (control). 

{\bf Contributions:} (1) We modify the standard policy gradient objective to include a direct variance estimator for learning policies that maximize the variance-penalized return. (2) We develop a multi-timescale actor-critic algorithm, by  deriving the gradient of the variance estimator in both the on-policy and the off-policy case. (3) We prove convergence to locally optimal policies in the on-policy tabular setting. (4) We compare our proposed variance-penalized actor-critic (\texttt{VPAC}) algorithm with two baselines: actor-critic \texttt{(AC)} \cite{sutton2000policy, konda2000actor}, and an existing \textit{indirect} variance penalized approach called variance-adjusted actor-critic (\texttt{VAAC}) \cite{tamar2013variance}. We evaluate our on- and off-policy \texttt{VPAC} algorithms in both discrete and  continuous domains. The empirical findings demonstrate that  \texttt{VPAC}  compares favorably to both baselines in terms of the mean return, but generates trajectories with significantly lower variance in the return.


\section{Preliminaries}
\paragraph{Notation} We consider an infinite-horizon discrete MDP $\langle \mathcal{S}, \mathcal{A}, \mathcal{R}, P, \gamma \rangle$ with finite state space $\mathcal{S}$ and finite action space $\mathcal{A}$. $\mathcal{R} \in \R$ denotes the reward function (with $R_{t+1}$ denoting the reward at time $t$). A policy $\pi: \mathcal{S}\rightarrow \mathcal{A}$ governs the behavior of the agent in state $s$, the agent chooses an action $a\sim \pi(\cdot|s)$, then transitions to next state $s'$ according to transition probability $P(s'|s,a)$. $\gamma \in [0,1]$ is the discount factor. Let $G_t = \sum_{l=0}^\infty \gamma^l R_{t+1+l}$ denote the accumulated discounted reward (also known as \textit{return}) along a trajectory. The state value function for $\pi$ is defined as: $V_\pi(s) = \E_{\pi}[G_t | S_t =s]$ and state-action value function is: $Q_\pi(s,a) = \E_{\pi}[G_t | S_t =s, A_t = a]$. In this paper, $\E_{\pi}[.]$ denotes expectation over transition function of MDP and probability distribution under $\pi$ policy.

\paragraph{Actor-Critic (AC) }
The policy gradient (PG) method \cite{sutton2000policy} is a policy optimization algorithm that performs gradient ascent in the direction maximizing the expected return. Given a parameterized policy $\pi_{\theta}(a| s)$, where $\theta$ is the policy parameter, an initial state distribution $d_{0}$ and the discounted weighting of states $d_{\pi}(s) = \sum_{t=0}^{\infty} \gamma^t {P(S_t = s | s_{0} \sim d_{0}, \pi)}$ encountered starting at some state $s_0$, the gradient of the objective function $J_{d_0}(\theta) = \sum_{s_0}d_{0}(s_0) V_{\pi_\theta}(s_0)$~\cite{sutton2018reinforcement} is given by: 
\begin{align}
\nabla_{\theta} J_{d_0}(\theta) = \E_{s_0 \sim d_0}\Big[\sum_{s} d_\pi (s) \sum_{a} \nabla_{\theta} \pi_\theta(a|s) Q_\pi(s,a)\Big]. \label{def:pgt-obj}
\end{align}

Actor-critic (AC) algorithms~\citep{sutton2000policy,konda2000actor} build on the PG theorem and learn both a policy, called the actor, and a value function, called the critic, whose role is to provide a good PG estimate. The one-step AC update of the policy parameter is given by:
\begin{align}
    \theta_{t+1} &= \theta_t + \alpha (G_{t:t+1} )\nabla_\theta \log \pi(A_t|S_t, \theta),
\end{align}
where $G_{t:t+1} = R_{t+1} + \gamma V(S_{t+1})$ is estimated by bootstrapping with the next state value function.

\section{On-Policy Variance-Penalized Actor-Critic (VPAC)}\label{sect:on_policy_vpac}
The modified objective function of our proposed approach is given as:

\begin{align}\label{def:vpac-obj}
 J_{d_0}(\theta) = \E_{s \sim d_0}\Big[\sum_{a}\pi_\theta(a|s)\Big(Q_\pi(s,a) - \psi \sigma_\pi(s,a)\Big)\Big],
\end{align}
where $d_0$ is an initial state distribution, $\sigma_{\pi}(s,a)$ is the variance in the return under the policy $\pi$, and $\psi \in [0, \infty)$ is the ``mean-variance'' trade-off parameter for the return. One can also recover the the conventional AC objective by simply setting $\psi=0$. For completeness, we present the derivation for $\sigma_{\pi}(s,a)$ in theorem \ref{thm:on-pol-var}.

\begin{definition}\label{def:sigma}
Given a state-action pair, variance in the return, $\sigma$, is defined as 
\begin{equation}
    \begin{split}
        &\sigma_\pi(s,a) =\\ 
        &\mathbb{E}_\pi\bigg[\Big(G_{t,\pi} - \mathbb{E}_\pi[G_{t,\pi}| S_t=s, A_t =a]\Big)^2 \Big| S_t=s, A_t =a\bigg].      
    \end{split}
\end{equation}
\end{definition}
Using Definition \ref{def:sigma}, the state variance function is denoted by $\sigma_\pi(s) = \sum_a \pi(a|s) \sigma_\pi(s,a)$. 
\begin{theorem}
\label{thm:on-pol-var}
(\textbf{On-policy variance in return}): Given a policy $\pi$ and $\bar\gamma = \gamma^2$, the variance in return for a state-action pair $\sigma_\pi(s,a)$ can be computed using Bellman equation as:
\begin{equation} \label{var_td}
\begin{split}
\sigma_\pi(s,a) & = \mathbb{E}_\pi\big[\delta_{t,\pi}^2  + \bar\gamma\sigma_\pi(S_{t+1}, A_{t+1})\big| S_t=s, A_t=a\big],\\
\text{where, }\\
\delta_{t,\pi} &= R_{t+1} + \gamma Q_\pi(S_{t+1},A_{t+1}) - Q_\pi(S_{t},A_{t}).
\end{split}
\end{equation}
Proof in Appendix A.
\end{theorem}

\begin{definition}\label{def:kstep}
The $\gamma$-discounted $k$-step transition is defined as
\begin{equation}\label{kstep_transition}
 P_\gamma^{(k)}(S_{t+k}|S_t) = P_\gamma^{(1)}(S_{t+k}|S_{t+k-1})\!\times\! P_\gamma^{(k-1)}(S_{t+k-1}|S_{t}),  
\end{equation}
where, 1-step transition is \[P_\gamma^{(1)}(S_{t+1}|S_t) = \gamma\sum_{a}\pi_\theta(a|S_t)P(S_{t+1}| S_t, a).\]
\end{definition}
\begin{theorem}
\label{thm:on-pol-pg}
(\textbf{Variance-penalized on-policy PG theorem}): Given $s_{0} \sim d_{0}$ - an initial state distribution,  a stochastic policy $\pi_\theta$, $\bar\gamma = \gamma^2$, the gradient of the objective $J$ in \eqref{def:vpac-obj} w.r.t. $\theta$ is given by
\begin{equation}
    \begin{split}
       \nabla_{\theta} J_{d_0}(\theta)=&\mathbb{E}_{s_0 \sim d_0}\Bigg[\sum_{k=0}^{\infty}\sum_{s}\Big[ \\
       &P_{\gamma}^{(k)}(s|s_0)\sum_{a}\nabla_\theta \pi_\theta(a|s) Q_\pi(s,a)\\
       &- \psi P_{\bar\gamma}^{(k)}(s|s_0)\sum_{a}\nabla_\theta \pi_\theta(a|s) \sigma_\pi(s,a)\Big]\Bigg].
    \end{split}
\end{equation}
Proof in Appendix A.
\end{theorem}
{
	\begin{algorithm}[h]
		\caption{On-policy VPAC (changes to AC in \textcolor{blue}{blue})}
		\label{alg:on_VPAC}
		\begin{algorithmic}[1]
			\STATE $\alpha_w, \alpha_\theta, \alpha_z$ stands for the step size of critic, policy and variance respectively.
			\STATE \textbf{Input}: differentiable policy $\pi_{\boldsymbol{\theta}}(a|s)$, value $\hat{Q}(s,a,\boldsymbol{w})$, and \textcolor{blue}{variance $\hat{\sigma}(s,a,\boldsymbol{z})$}
			\STATE \textbf{Parameters}: $\gamma \in [0,1]$, $\psi \in [0, \infty)$, $\alpha_\theta \textless \alpha_z  \textless \alpha_w $, $\textcolor{blue}{\bar{\gamma}=\gamma^2}$
			\STATE Initialize parameters $\theta, w, z$
			\FOR{Episode $i = 1,2, \dots$}
    			\STATE Initialize $S$, sample $ A\sim \pi_\theta(.|S)$
    			\STATE $I_Q, \textcolor{blue}{I_{\sigma}} = 1, \textcolor{blue}{1}$
    			\REPEAT
    			\STATE Take action $A$, observe $\{R, S'\}$; sample $A' \sim \pi_\theta(.|S')$
    			\STATE $\delta \gets R + \gamma\hat{Q}(S',A',w) - \hat{Q}(S,A, w)$
    			\STATE $\textcolor{blue}{\bar{\delta} \gets \delta^2 + \bar{\gamma}\hat{\sigma}(S',A',z) - \hat{\sigma}(S,A, z)}$
    			\STATE $w \gets w +\alpha_w \delta \nabla_w \hat{Q}(S,A,w)$
    			\STATE $\textcolor{blue}{z \gets z +\alpha_z \bar{\delta} \nabla_z \hat{\sigma}(S,A,z)}$
    			\STATE $\theta \gets \theta +\alpha_\theta \nabla_\theta \log(\pi_\theta(A|S))\Big($
    			\STATE $\quad \quad \quad I_{Q}\hat{Q}(S,A,w) \textcolor{blue}{- \psi I_{\sigma}\hat{\sigma}(S,A,z)}\Big)$
    			\STATE $I_{Q}, \textcolor{blue}{I_{\sigma}} = \gamma I_{Q}, \textcolor{blue}{\bar\gamma I_{\sigma}}$
    			\STATE $S \gets S', A \gets A'$
                
    			\UNTIL{$S'$ is a terminal state}
			\ENDFOR
		\end{algorithmic}
	\end{algorithm}
}
Algorithm \ref{alg:on_VPAC} contains the pseudo-code of our proposed method. For a finite-state, discrete MDP, our algorithm can be shown to converge to a locally optimal policy using the ordinary differential equation (ODE) approach  used in stochastic approximation~\cite{borkar2009stochastic} (See Appendix C for the complete proof). 
To ensure convergence of our method, the step size parameters are selected such that the 
$Q(\cdot)$ is the first to converge followed by $\sigma(\cdot)$ and $\pi(\cdot)$. An immediate consequence is that the value estimate $\hat Q$  will almost converge to $Q_\pi$ before $\sigma$ is updated, so we can use $\hat Q$ for computing $\delta_{t,\pi}^2$ \eqref{var_td}. Further, \citeauthor{sherstan2018directly} theoretically showed that if the value function does not satisfy the Bellman operator of the expected return, the error in estimation of variance using the above formulation is proportional to the error in the value function estimate. The theoretical analysis of a biased value function on the performance of variance-penalized policy is an interesting direction and is left for the future work.

\section{Off-Policy Variance-Penalized Actor-Critic (VPAC)}
In the off policy setting, the experience generated by a behaviour policy $b$ is used to learn a target policy $\pi$. We modify the objective function as in \eqref{obj_offp} to include an importance sampling correction factor $\rho(s,a) = \frac{\pi(a|s)}{b(a|s)}$ which accounts  for the discrepancy between the policy distributions:
\begin{align}\label{obj_offp}
J_{d_0}(\theta) = \mathbb{E}_{s \sim d_0,a \sim b}\Big[\rho(s,a)(Q_\pi(s, a) - \psi \sigma_\pi(s,a))\Big].
\end{align}
\begin{definition}
Return given a state-action pair under a target policy $\pi$ when actions are sampled from a behavior policy $b$ is:
\begin{equation}
    \begin{split}
        G_{t,\pi,b} &= R_{t+1} +  \gamma \rho_{t+1} G_{t+1,\pi,b}. 
    \end{split}
\end{equation}
\end{definition}
$R_{t+1}$ does not have a correction factor, because, $G_{t,\pi,b}$ is described given a state-action pair (action $A_t$ is already given). We extend the off-policy variance given a state $\sigma(s)$ \cite{sherstan2018directly} to a state-action pair $\sigma(s,a)$. We re-write the Bellman equation derivation for the off-policy variance (Theorem \ref{thm:sigma_th_offpac}) for completeness.
\begin{theorem}
\label{thm:sigma_th_offpac}
(\textbf{Off-policy variance in return}): 
Given a behaviour policy $b$ and $\bar\gamma = \gamma^2$, the off-policy variance in return for a state-action pair $\sigma_\pi(s,a)$ can be computed using Bellman equation as follows:
\begin{equation}
\begin{split}
&\sigma_\pi(s,a) =\\
&\mathbb{E}_b\big[\delta_{t,\pi}^2 + {\bar\gamma}\rho_{t+1}^2 \sigma_\pi(S_{t+1}, A_{t+1})\big| S_t=s,A_t=a\big],\label{variance_offp}
\end{split}
\end{equation}
where,
\begin{equation} \label{var_td_off_policy}
    \begin{split}
        \delta_{t,\pi} = & R_{t+1} + \gamma \rho_{t+1} Q_{\pi}(S_{t+1}, A_{t+1})\\
        &- Q_\pi(S_{t}, A_{t}).
    \end{split}
\end{equation}
Proof in Appendix B.
\end{theorem}
The above theorem provides a method to relate the  variance under a target policy from the current to the next state-action pair when the trajectories are generated from a different behaviour policy. Here, $\mathbb{E}_b[.]$ denotes the expectation over the transition function and actions drawn from a behavior policy distribution.
\begin{definition}
The $\gamma$-discounted $1$-step transition under a target-policy $\pi$ is
\begin{equation}\label{transition_off_p_Q}
\begin{split}
   T_{\gamma}^{(1)}(S_{t+1},A_{t+1}|S_{t},A_{t}) = &\gamma \rho_{t+1} P(S_{t+1}|S_{t},A_t)\\
   & \times b(A_{t+1}|S_{t+1}). 
\end{split}
\end{equation}
\end{definition}
\begin{definition}
Let $\bar T$ be 1-step $\bar \gamma$-discounted transition
\begin{equation}\label{transition_var_off_pol}
\begin{split}
     \bar{T}_{\bar\gamma}^{(1)}(S_{t+1},A_{t+1}| S_t,A_t) = &\bar\gamma \rho_{t+1}^2 P(S_{t+1}|S_t,A_t)\\
     & \times b(A_{t+1}|S_{t+1}). 
\end{split}
\end{equation}
\end{definition}
One can also define the $k$-step transition here similar to Definition \ref{def:kstep}. The off-policy state-action value $Q_\pi(s,a)$ using \eqref{transition_off_p_Q} with importance sampling correction factor included is defined as:
\begin{equation}\label{Q_off_pol}
    Q_\pi(s,a) = r(s,a) + \sum_{s', a'}T_{\gamma}^{(1)}(s',a'|s,a) Q_\pi(s',a').
\end{equation}
\begin{theorem}
\label{thm:Jofpac_paper}
(\textbf{Variance-penalized off-policy PG theorem}): Given $s_0 \sim d_0$, $a_0 \sim b$, a stochastic target policy $\pi_\theta$, $T_\gamma^{(k)}(s,a|s_0,a_0)$ and $\bar{T}_{\bar\gamma}^{(k)}(s,a|s_0,a_0)$ following \eqref{transition_off_p_Q} and \eqref{transition_var_off_pol} respectively, the gradient of the objective function $J$ in \eqref{obj_offp} w.r.t. $\theta$ is 
\begin{equation*}
\begin{split}
    &\nabla_{\theta}J_{d_{0}}(\theta) = \mathbbm{E}_{s_0 \sim d_0, a_0 \sim b}\Bigg[\sum_{k=0}^{\infty} \sum_{s,a} \Big[\\
    &T_\gamma^{(k)}(s,a|s_0,a_0)\nabla_\theta\log\pi_\theta(a|s)Q_\pi(s,a)\\
    &- \psi \bar{T}_{\bar\gamma}^{(k)}(s,a|s_0,a_0)
        [1 + \mathbbm{1}_{k\geq 1}]\nabla_\theta \log\pi_\theta(a|s)\sigma_\pi(s,a)\Big]\Bigg].
\end{split}
\end{equation*}
Proof in Appendix B.
\end{theorem}
Here, the importance sampling factor is rolled inside the $T$ and $\bar T$ terms. This is an incremental update which uses the experience from the exploratory behavior policy to improve a different target policy. Similar to on-policy \texttt{VPAC}, using the  multi-timescale argument, we use the value estimate $\hat Q$, instead of the true state-action value $Q_\pi$ in the calculation of $\delta_{t,\pi}^2$ in \eqref{var_td_off_policy}. Algorithm 1 in Appendix D shows a prototype implementation for off-policy \texttt{VPAC}.

\begin{figure*}[h]
		\begin{subfigure}[b]{0.3\textwidth}
			\centering
			\captionsetup{justification=centering}
			\includegraphics[width=0.48\linewidth]{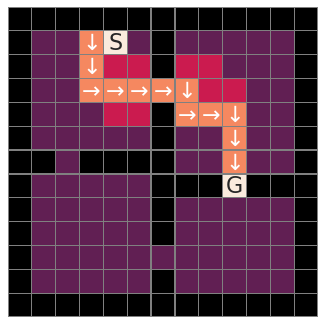}
			\caption[]{\small{AC trajectory}}
		\end{subfigure}
		\begin{subfigure}[b]{0.3\textwidth}
			\centering
			\captionsetup{justification=centering}
			\includegraphics[width=0.6\linewidth]{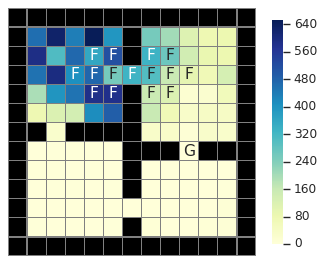}
			\caption[]{\small{AC variance}}
		\end{subfigure}
		\begin{subfigure}[b]{0.3\textwidth}
			\centering
			\captionsetup{justification=centering}
			\includegraphics[width=0.6\linewidth]{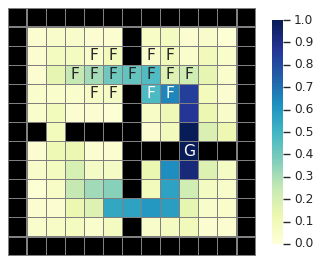}
			\caption[]{\small{AC frequency}}
		\end{subfigure}
		\newline
		\begin{subfigure}[b]{0.3\textwidth}
			\centering
			\captionsetup{justification=centering}
			\includegraphics[width=0.48\linewidth]{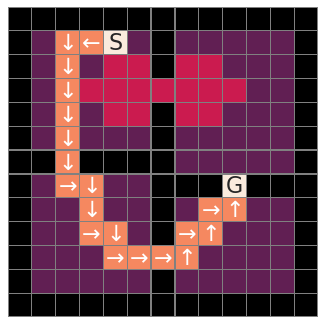}
			\caption[]{\small{VPAC trajectory}}
		\end{subfigure}
		\begin{subfigure}[b]{0.3\textwidth}
			\centering
			\captionsetup{justification=centering}
			\includegraphics[width=0.6\linewidth]{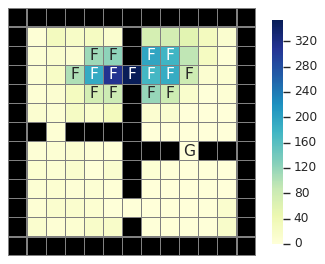}
			\caption[]{\small{VPAC variance}}
		\end{subfigure}
		\begin{subfigure}[b]{0.3\textwidth}
			\centering
			\captionsetup{justification=centering}
			\includegraphics[width=0.6\linewidth]{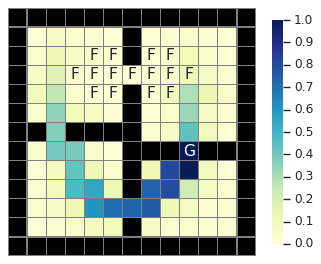}
			\caption[]{\small{VPAC frequency}}
		\end{subfigure}
		\caption[]
        {\textbf{Visualizations in four rooms}: Qualitative analysis of converged policy's behavior in  \texttt{AC} and \texttt{VPAC} algorithms. \texttt{AC} algorithm:  (a),(b),(c) Sampled trajectory, variance in return for initial state distribution, state visitation frequency over $10000$ trajectories respectively. Similarly, (d),(e),(f) depicts outcome on \texttt{VPAC} algorithm. $F$/ red region depicts the frozen states, $S$ the start state and $G$ the goal state. \texttt{VPAC} produces a lower variance in the return than \texttt{AC} by taking lower hallway to avoid the variable $F$ region. } 
        \label{fig:policy_fourroom}
\end{figure*}    
\section{Experiments}
We present an empirical analysis in both discrete and continuous environments for the proposed on-policy and off-policy \texttt{VPAC} algorithms. We compare our algorithms with two baselines: \texttt{AC} and \texttt{VAAC}, an existing variance penalized actor-critic algorithm using an \textit{indirect} variance estimator \cite{tamar2013variance} (refer Related Work section for further details). Given the penalty term is added to the objective function, we hypothesize that our approach should achieve a reduction in variance, but on-par average returns compared to the baselines. Implementation details along with the hyperparameters used for all the experiments\footnote{\textbf{Code} for all the experiments is available on \url{https://github.com/arushi12130/VariancePenalizedActorCritic.git}} are provided in Appendix E.



\subsection{On-Policy Variance Penalized Actor-Critic (VPAC)}
\label{sec:onpolicy_experiments}
\subsubsection{Tabular environment}
We modify the classic four rooms (FR) environment \cite{sutton1999between} to include a patch of frozen states (see Fig.~\ref{fig:policy_fourroom}) with stochastic reward. In the normal (non-frozen) states, the agent gets a reward of $0$, whereas in the frozen states, the reward is sampled from a normal distribution $\mathcal{N}(\mu=0, \sigma=8)$. Upon reaching the goal, a reward of $50$ is observed. Note that in expectation, the reward for the normal and the frozen states is the same. Hence, an agent that only optimizes expected return would have no reason to prefer some of these states over others. However, intuitively it would make sense to design agents that avoid the frozen states, which can be achieved by algorithms sensitive to the variance. We keep $\gamma=0.99$. We use Boltzmann exploration and do a grid search to find the best hyperparameters for all algorithms, where the least variance is used to break ties among policies with maximum mean performance. We show the impact of varying the hyperparameters on both the mean and the variance performance in Appendix E. We also show the table of best hyperparameters (found using grid search) used for all algorithms in Appendix E.

\begin{figure*}[h]
    \begin{center}
        \includegraphics[width=0.8\textwidth]{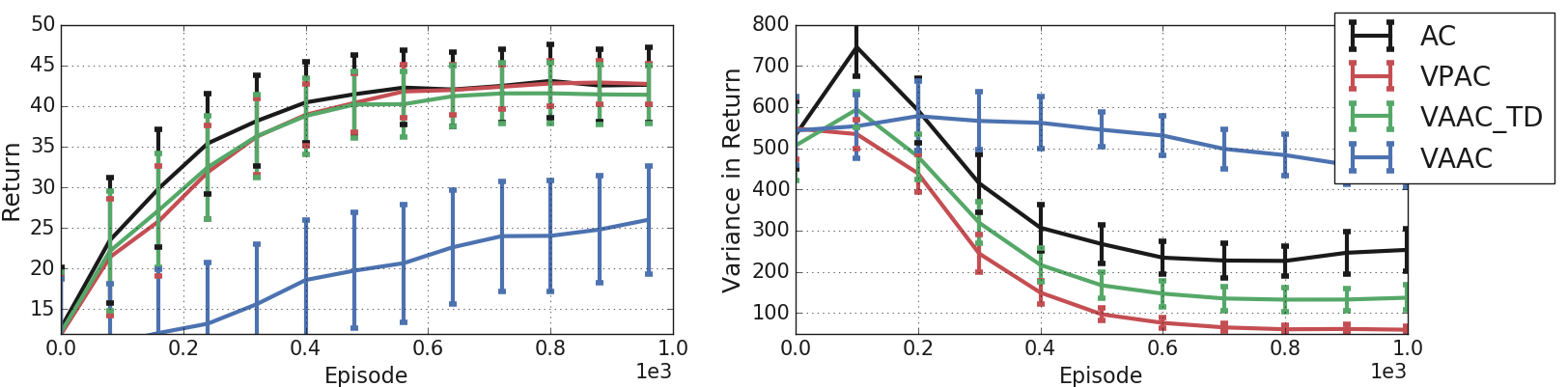}
        \caption{\textbf{Performance in four rooms}: Mean and variance performance comparison of our \texttt{VPAC} and baselines ( \texttt{AC, VAAC, VAAC\_TD}) algorithms where standard deviation error-bars from $100$ runs are shown. \texttt{VPAC} achieves significantly lower variance in the return (right plot) while maintaining similar average return as  the baselines (left plot).}
    \label{fig:FourRoomAC}
    \end{center}
\end{figure*}
\begin{figure}[h]
		\begin{center} \includegraphics[width=0.95\linewidth]{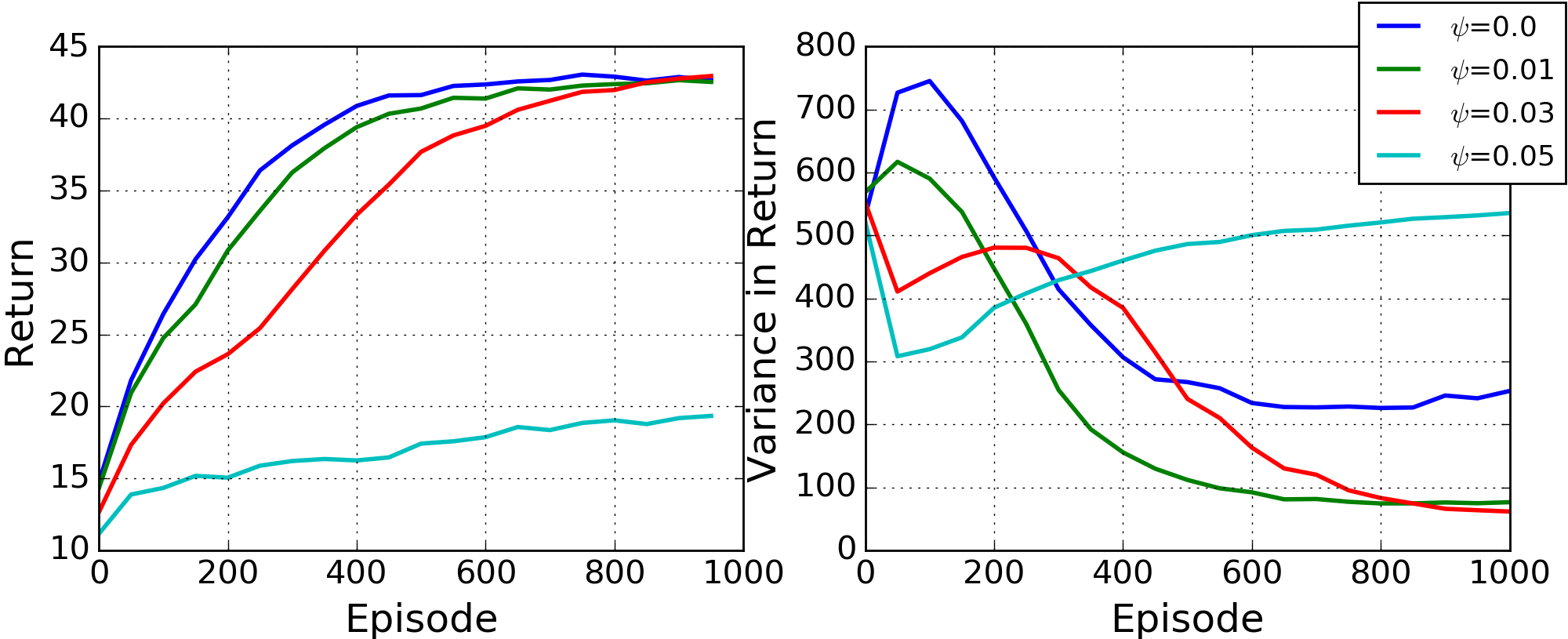}
        \caption[]
        {\textbf{Mean-variance tradeoff ($\psi$) vs performance in four rooms}: Plot shows \texttt{VPAC}'s learning curves for different values of $\psi$ in terms of the mean and the variance performance. The performance is averaged over $100$ runs.} 
        \label{fig:fr_comparison_psi}
        \end{center}
\end{figure}

In Fig.~\ref{fig:FourRoomAC}, we compare the mean and the variance in the return of the proposed Algorithm~\ref{alg:on_VPAC} \texttt{VPAC} and two other indirect variance penalized algorithms: Algorithm 2 \texttt{VAAC} \cite{tamar2013variance} (a Monte-Carlo critic update), and Algorithm 3 \texttt{VAAC\_TD} (we modified  \texttt{VAAC} to do a TD critic update for a  fair comparison) and risk-neutral \texttt{AC}. Algorithms 2 and 3 are presented in Appendix E. The figure shows that \texttt{VPAC} has comparable mean performance to \texttt{AC} and less variance in return (red line). For all algorithms, we rolled out $800$ trajectories for each policy along the learning curve to calculate the variance in return. Note, here variance penalization does not highly impact the mean performance even though number of steps to reach the goal increases, because $\gamma = 0.99$. In Fig. \ref{fig:fr_comparison_psi} we show the effect of varying the mean-variance tradeoff $\psi$ on both the mean and the variance performance of \texttt{VPAC} algorithm. This shows with very high values of $\psi$, the exploration is curbed causing a decay in the performance. In Appendix E, we show the sensitivity of \texttt{VPAC} and \texttt{VAAC\_TD} with the step size ratios of policy, variance and value function. We empirically observe that keeping the step sizes of value and variance function closer to each other and the step size of policy very small in comparison to the other two, results in a better performance (both higher mean and lower variance). 

\begin{figure}[h]
		\begin{center} \includegraphics[width=0.95\linewidth]{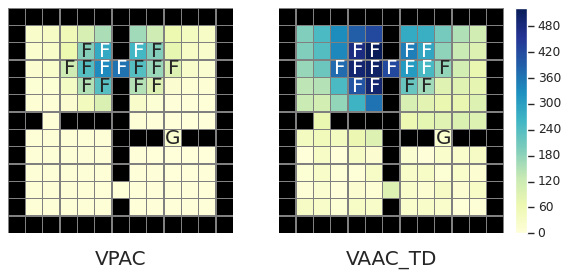}
        \caption[]
        {\textbf{Comparison of variance in return for four rooms}: Compares the converged policy's variance in return for initial state distribution of \texttt{VPAC} and baseline \texttt{VAAC\_TD} algorithm. The variance in performance over $5000$ trajectories is averaged over 50 runs. \texttt{VPAC}'s converged policy achieves significantly lower variance in return demonstrating the effectiveness in avoiding the variable $F$ region.}
        \label{fig:fr_comparison_variance_vpac_vaactd}
        \end{center}
\end{figure}

\textit{Qualitative analysis: }
We compare \texttt{AC} and \texttt{VPAC} to analyse the learned policy's behaviour. Fig. \ref{fig:policy_fourroom}, (a) \& (d), shows the sampled trajectory where \texttt{VPAC} clearly learns to avoid the variance-inducing  \textit{frozen} region. The variance in the return is depicted in Fig. \ref{fig:policy_fourroom}, (b) \& (e), where each cell color represents variance intensity in trajectories initialized from that cell. \texttt{VPAC} shows smaller variance compared to \texttt{AC}, and its trajectories avoid the $F$ region. This  is further strengthened by Fig. \ref{fig:policy_fourroom}, (c) \& (f), showing the state visitation frequency, where \texttt{VPAC} has higher visitation frequency for the lower hallway. Fig. \ref{fig:fr_comparison_variance_vpac_vaactd} shows the comparison between \texttt{VPAC} and \texttt{VAAC\_TD} algorithm's observed variance in return for a converged policy (obtained after $1000$ episodes) being initialized from each different states in the FR environment. Policy learnt using \texttt{VPAC} algorithm displays lower variance, highlighting better capability in learning and avoiding states that cause variability in the performance.

\subsubsection{Continuous state-action environments}
We now turn to continuous state-action tasks in the MuJoCo OpenAI Gym \citep{brockman2016openai}. Note, here our aim is not to beat the mean performance with state-of-the-art PG methods, but to show the effect of introducing variance-penalization in a PG method. We can use the proposed objective with various existing PG methods, here, we limit our comparison with the proximal policy optimization (PPO) algorithm \citep{schulman2017proximal}.

\begin{figure}[h]
        \centering
		\begin{subfigure}[b]{0.45\linewidth}
			\centering
			\captionsetup{justification=centering}
			\includegraphics[width=0.8\textwidth]{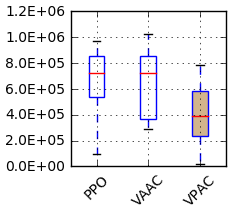}
			\caption[]{\small{Hopper}}
		\end{subfigure}
		\begin{subfigure}[b]{0.45\linewidth}
			\centering
			\captionsetup{justification=centering}
			\includegraphics[width=0.8\textwidth]{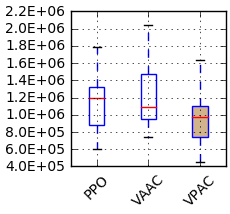}
			\caption[]{\small{Walker2D}}
		\end{subfigure}
		
		\begin{subfigure}[b]{0.45\linewidth}
			\centering
			\captionsetup{justification=centering}
			\includegraphics[width=0.8\textwidth]{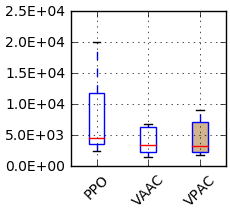}
			\caption[]{\small{HalfCheetah}}
		\end{subfigure}
		\caption[]{\textbf{Variance analysis in Mujoco}: The box plots show the distribution of variance in the  return of converged policy across multiple runs for \texttt{VPAC} and baselines \texttt{PPO} and \texttt{VAAC}.  Overall, in comparison to the baselines, \texttt{VPAC} exhibits a lowered, tighter distribution along with a reduction in the median of variance.} 
        \label{fig:mujoco_boxplot}
\end{figure}

A separate network for variance estimation is added in \texttt{PPO} to implement the variance-penalized objective function. Further details are provided in Appendix E. We compare the performance of our method \texttt{VPAC} with standard \texttt{PPO} and self-implemented \texttt{VAAC} in the \texttt{PPO} framework. Fig.~\ref{fig:mujoco_boxplot} shows the box plot of the variance in the return for converged policies across multiple runs in Hopper, Walker2D, and HalfCheetah environments. As seen in the figure, a lowered concentration mass for the variance distribution is observed for \texttt{VPAC} in comparison to the baselines, supporting a reduction in the variance induced by the algorithm (see red median line and inter-quartile range). Table~\ref{tab:Mojoco} shows the mean and the variance in return from $100$ rolled out trajectories of the converged policies of different algorithms. We averaged the above performance measure over multiple runs. Our method \texttt{VPAC} observes a reduction in the variance, but also suffers slightly in terms of mean performance. Note, Table~\ref{tab:Mojoco} and Fig.~\ref{fig:mujoco_boxplot} shows different metrics, mean and median of variance in performance respectively over multiple runs. The learning curve mean performance for different algorithms is shown in Appendix E.

\begin{table*}[t]
  \caption{\textbf{Performance in Mujoco}: Compares the averaged performance of \texttt{PPO,VAAC,VPAC} algorithms over multiple runs in terms of the mean and the variance in the score over $100$ trajectories. Bold highlights the least variance in the score. Numbers in braces show the percentage reduction of variance in comparison to \texttt{PPO}. \texttt{VPAC} achieves a lower variance in the score compared to the baselines, but also suffers slightly in terms of mean performance.}
  \label{tab:Mojoco}
  \centering
  \begin{tabular}{lllllll}
    \toprule
    {} & \multicolumn{2}{c}{\textbf{PPO}}  & \multicolumn{2}{c}{\textbf{VAAC}} & \multicolumn{2}{c}{\textbf{VPAC (ours)}}\\
    
    \cmidrule{2-3}
    \cmidrule{4-5}
    \cmidrule{6-7}
    Environment & Mean & Var (1e5) & Mean & Var (1e5) & Mean & Var (1e5)\\
    \midrule
    HalfCheetah & 1557 & {1.6} & {1525} & 0.8 (50\%) & {1373} & \textbf{{0.1}} (93\%)              \\
    Hopper & {1944} & {6.6} & 1991 & {6.5} (1.5\%) & 1624 & \textbf{{4.0}} (39.4\%)          \\
    Walker2d & {3058}  & {12.1}  & {3102}  & {12.5} (-3.3\%) & {2625}  & \textbf{{9.2}} (23.9\%) \\
    \bottomrule
  \end{tabular}
\end{table*}


\begin{figure*}[!ht]
		\begin{center}
        \begin{subfigure}[b]{0.48\textwidth}
        \centering
             \includegraphics[width=\linewidth]{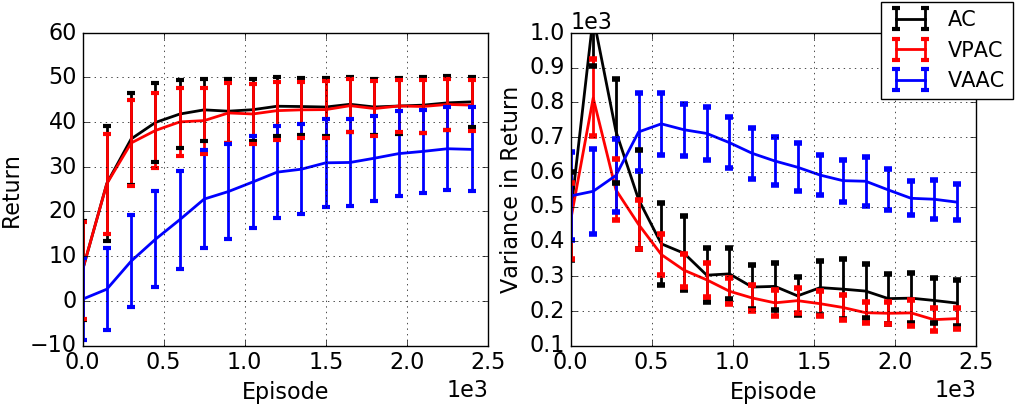}
            \caption[]%
            {{\small Discrete puddle-world}}   
            \label{fig:PuddleDiscrete}
        \end{subfigure}
        \quad
        \begin{subfigure}[b]{0.48\textwidth}
        \centering
             \includegraphics[width=\linewidth]{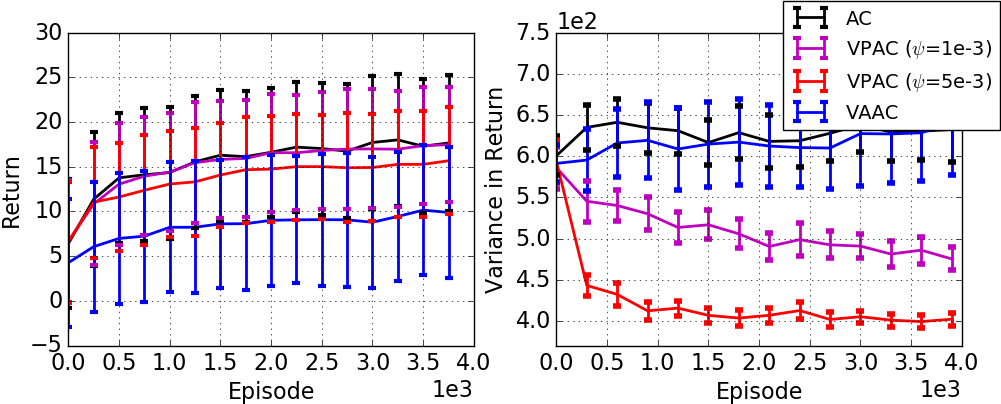}
            \caption[]%
            {{\small Continuous puddle-world}}    
            \label{fig:PuddleContinuous}
        \end{subfigure}
        \caption{\textbf{Off-policy learning curves}: \textit{Off-policy} performance comparison over baselines \texttt{AC, VAAC} and \texttt{VPAC} averaged over $100$ trials a) discrete, and $50$ trials for b) continuous setting in puddle-world environment. The graph shows both the mean and the variance in return for the learnt target policy. $\psi$ is mean-variance tradeoff.}
    \label{fig:OptimalPuddle_offpolicy}
    \end{center} 
\end{figure*}

\subsection{Off-Policy Variance-Penalized Actor-Critic (VPAC)}
\label{sec:offpolicy_experiments}
We compare off-policy \texttt{VPAC} with both \texttt{VAAC}~\cite{tamar2013variance} and \texttt{AC} algorithms. Since \texttt{VAAC} is a on-policy AC algorithm, we modify it to its off-policy counterpart by appropriately incorporating an importance sampling correction factor. Note that, \texttt{VAAC} uses Monte-Carlo critic, whereas, \texttt{AC} and \texttt{VPAC} use TD critic. We left the comparison with off-policy version of \texttt{VAAC\_TD}, since its derivation in AC was not straightforward.

\subsubsection{Tabular environment}
\begin{figure}[th]
		\begin{center}
        \begin{subfigure}[b]{0.22\textwidth}
        \centering
             \includegraphics[width=0.6\textwidth]{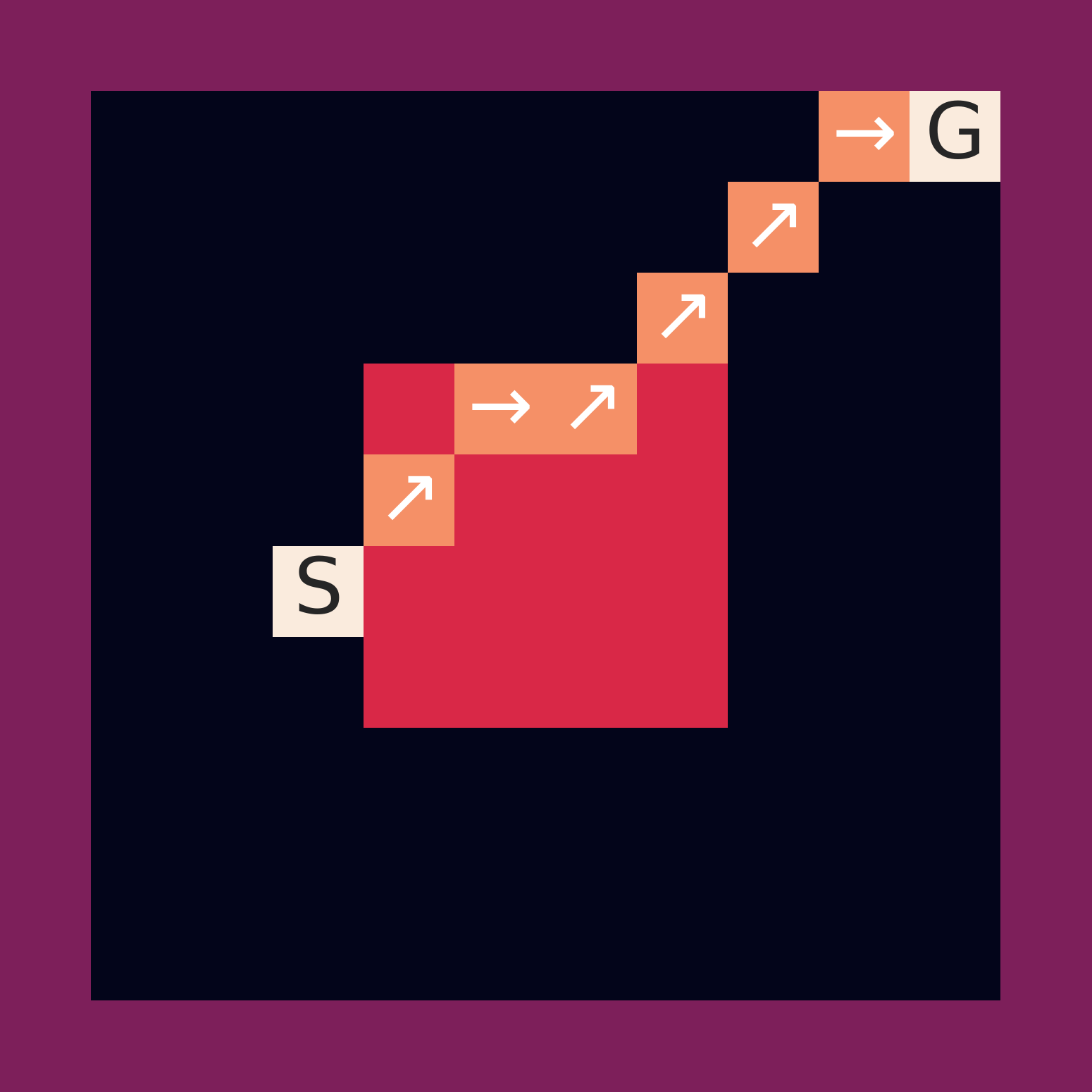}
            \caption[]%
            {{\small Discrete AC}}   
            \label{fig:PuddleDiscreteTrajUS}
        \end{subfigure}
        \begin{subfigure}[b]{0.22\textwidth}
        \centering
             \includegraphics[width=0.6\textwidth]{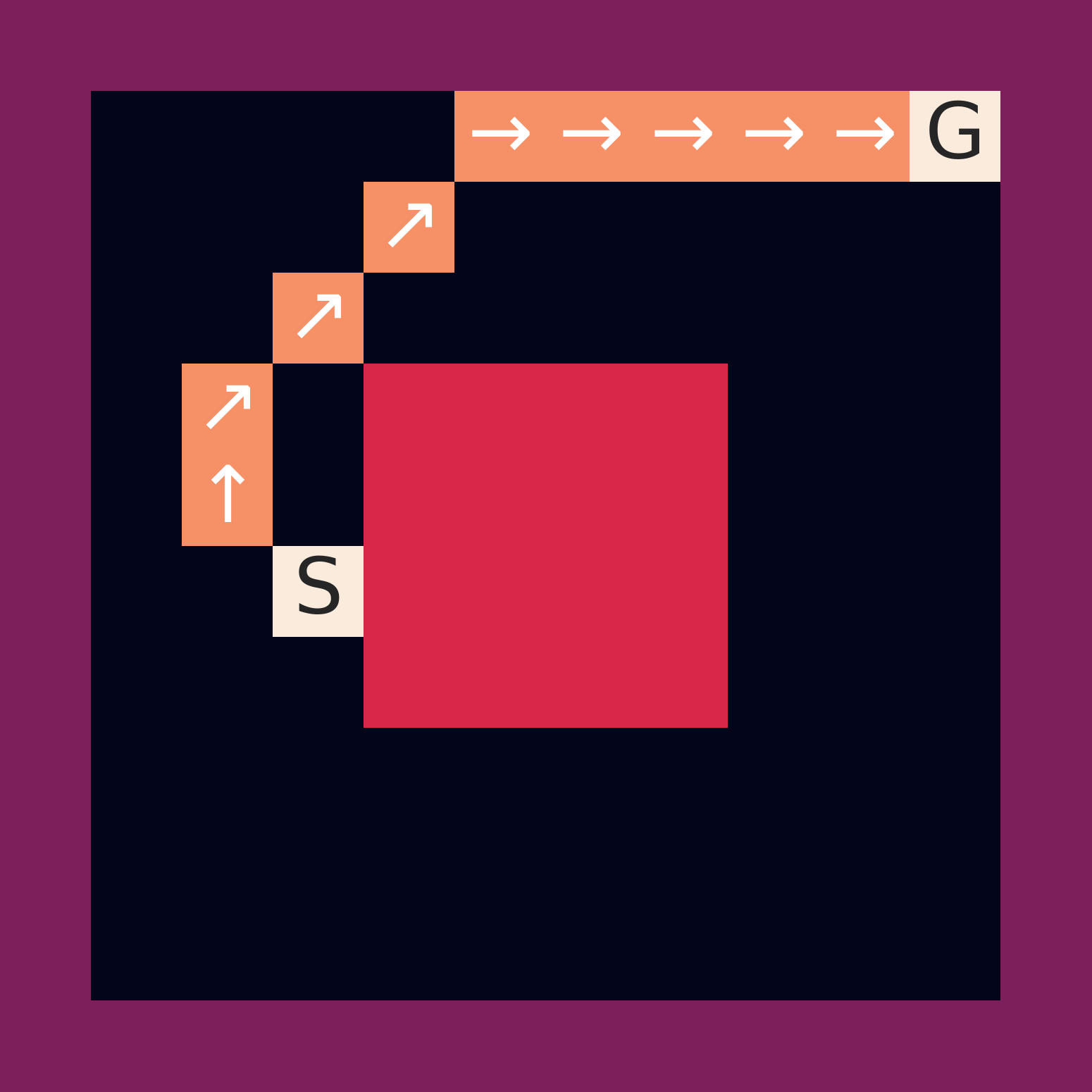}
            \caption[]%
            {{\small Discrete VPAC}}    
            \label{fig:PuddleDiscreteTrajS}
        \end{subfigure}
        \begin{subfigure}[b]{0.22\textwidth}
        \centering
             \includegraphics[width=0.62\textwidth]{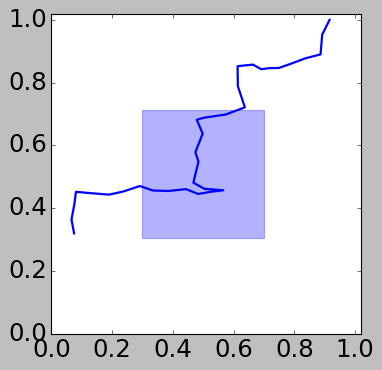}
            \caption[]%
            {{\small Continuous AC}}    
            \label{fig:PuddleContTrajUS}
        \end{subfigure}
        \begin{subfigure}[b]{0.22\textwidth}
        \centering
             \includegraphics[width=0.62\textwidth]{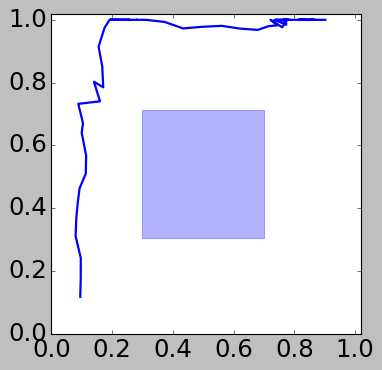}
            \caption[]%
            {{\small Continuous VPAC}}    
            \label{fig:PuddleContTrajS}
        \end{subfigure}
        \caption{\textbf{Off-policy puddle-world trajectories}: Compares sampled trajectories for both off-policy \texttt{AC} and \texttt{VPAC} in \{(a),(b)\} discrete, and \{(c),(d)\} continuous puddle-world environment. \texttt{VPAC} avoids variable reward puddle region (shown in red/blue).}
    \label{fig:puddle_traj}
    \end{center}
\end{figure}

We investigate a modified puddle-world environment with a variable reward puddle region in the centre. The goal state $G$ is placed in the top right corner of the grid. The reward from normal and puddle region is same on expectation. Samples are generated from a uniform behavior policy. The target policy is a Boltzmann distribution over policy parameters. We use Retrace \cite{munos2016safe} for off-policy correction. Fig.~\ref{fig:PuddleDiscrete} compares the mean and the  variance in return of target policy for off-policy baselines \texttt{AC,VAAC} with  \texttt{VPAC}. \texttt{VPAC} observes the least variance in the return, without sacrificing the mean performance. Fig.~\ref{fig:puddle_traj} (a),(b) compares a sampled trajectory for \texttt{AC} and \texttt{VPAC}. The baseline takes the shortest path to the goal, whereas, \texttt{VPAC} avoids the variable reward puddle region. The state visitation frequency plots are provided in Appendix E.

\subsubsection{Continuous state environment} Next, we examine the performance of off-policy \texttt{VPAC} in continuous 2-D puddle-world environment similar to the discrete case, but with linear function approximation. We use tile coding \citep{sutton1998reinforcement} for discretization of the state-space. Further experimental details are presented in Appendix E. Fig.~\ref{fig:PuddleContinuous}, compares the mean and the variance performance of \texttt{VPAC} with the baselines. We observe a significant reduction in the variance for \texttt{VPAC} as compared to other baselines. In Fig.~\ref{fig:PuddleContinuous}, as mean-variance tradeoff $\psi$ increases (compare pink and red lines), we observe not only substantially lower variance (right plot) for \texttt{VPAC}, but, also a slight reduction in the mean performance (left plot). This observation highlights that role of $\psi$ in the mean-variance tradeoff. Fig.~\ref{fig:puddle_traj} (c), (d) displays the sampled trajectories for \texttt{AC} and \texttt{VPAC} converged policies, highlighting \texttt{VPAC} avoids the variance inducing regions.

\section{Related Work}\label{sec:related_work}
There has been significant effort to reduce the variability in the performance of AI agents, by minimizing the \textit{risk} in the form of exponential utility function  \cite{howard1972risk,borkar2001sensitivity,basu2008learning,nass2019entropic}, Sharpe ratio \cite{sharpe1994sharpe}, worst-case outcome \cite{heger1994consideration,mihatsch2002risk}, value-at-risk (VaR) \cite{duffie1997overview}, conditional-value-at-risk (CVaR) \cite{rockafellar2000optimization,chow2014algorithms,tamar2015optimizing}, or variance \cite{markowitz1959portfolio, filar1989variance, sobel1982variance, white1994mathematical}. \citeauthor{garcia2015comprehensive} \shortcite{garcia2015comprehensive} provide a detailed analysis of different approaches to limit the variability. Although, many risk-sensitive RL methods have been introduced, variance based risk methods have an explicit advantage in being more interpretable \cite{markowitz2000mean}.

Our work involves mean-variance optimization, and therefore, we limit our discussion to variance-related risk measures. \citeauthor{sobel1982variance} \shortcite{sobel1982variance} introduced an indirect approach to estimate the variance using the first and the second moments of return. Later, \citeauthor{white2016greedy} \shortcite{white2016greedy} extended the traditional indirect variance estimator to $\lambda$-returns. Indirect approaches to estimate the variance have been studied ~\cite{tamar2012policy, tamar2013variance, tamar2013temporal, prashanth2013actor, prashanth2016variance} in many mean-variance optimization problems. \citeauthor{guo2012mean} \shortcite{guo2012mean} studied mean-variance optimization, with the aim to minimize the variance, assuming access to already optimal expected reward, dealing with a much simpler problem than ours. In the episodic AC setting, \citeauthor{tamar2013variance} \shortcite{tamar2013variance} studied a variance-penalized method with an indirect approach to estimate the variance (using the second moment of return). \citeauthor{prashanth2013actor} \shortcite{prashanth2013actor,prashanth2016variance} extended the indirect variance estimator to AC algorithm using simultaneous perturbation methods. Another risk-averse policy gradient method proposed by \citeauthor{bisi2019risk} \shortcite{bisi2019risk} measures risk using a different metric called reward volatility which captures the variance of the reward at each time step (as opposed to variance in return which measures variance in the accumulated rewards among trajectories). Reward constrained policy optimization (RCPO) \cite{tessler2018reward} approach uses a fixed constraint signal as a penalty in the reward function. In our work, we use varying variance in return as a constraint which depends on both the policy and $Q$ function, thus, making combined Bellman update for critic (value and variance) impossible unlike RCPO.

\citeauthor{thomas2015high} \shortcite{thomas2015high} estimated safety in the off-policy setting by bounding the probability of performance for a given confidence. \citeauthor{white2016greedy} \shortcite{white2016greedy}, \citeauthor{sherstan2018directly} \shortcite{sherstan2018directly} measured the variance for a target policy using off-policy samples in the policy evaluation setting. It is to be noted that an alternative approach to estimate the variance in return is by using the distribution over return \cite{Morimurareturn, bellemare2017distributional}.

Our work is closely related to \citeauthor{tamar2013variance} \shortcite{tamar2013variance}, \citeauthor{prashanth2013actor} \shortcite{prashanth2013actor} and \citeauthor{sherstan2018directly} \shortcite{sherstan2018directly}. The former \cite{tamar2013variance, prashanth2013actor} developed an AC method for mean-variance optimization using the return's first and second moments for variance estimation (an indirect variance computation approach described in Introduction). On the other hand, direct variance estimator proposed by \citeauthor{sherstan2018directly} for \textit{policy evaluation} provides an alternative over indirect estimators. Our work is an extension to both of these methods, wherein, we propose a new AC algorithm which uses the TD style direct variance estimator \cite{sherstan2018directly} to compute a variance-penalized objective function in a \textit{control setting}.

\section{Conclusion \& Future Work}
We proposed an on- and off-policy actor-critic algorithm for variance penalized objective which leverages multi-timescale stochastic approximations, where both value and variance critics are estimated in TD style. We use the \textit{direct} variance estimator for our proposed objective function. The empirical evidence in our work (see section Experiments) demonstrates that both our algorithms result in trajectories with much lower variance as compared to the risk-neutral and existing \textit{indirect} variance-penalized counterparts.

Furthermore, we provided convergence guarantees for the proposed algorithm in the tabular case. Extending theoretical analysis to the linear function-approximation is a promising direction for future work. Another potential direction is to study the effects of a scheduler on the mean-variance tradeoff $\psi$, which can provide a balance between exploration at beginning and variance reduction towards the later stage.

\section{Acknowledgments}
The authors would like to thank Pierre-Luc Bacon, Emmanuel Bengio, Romain Laroche and anonymous AAAI reviewers for the valuable feedback on this paper draft.

\end{document}


\maketitle
\appendix

\begin{lemma}\label{off_policy_cross_term}
$\mathbb{E}_b[\gamma \delta_{t,\pi,b} \rho_{t+1} \big(G_{t+1,\pi,b} -  Q_\pi(S_{t+1}, A_{t+1})\big) \big| S_t = s, A_t=a]  = 0$.
\end{lemma}
\begin{proof}
\begin{align}\label{delta_def}
\delta_{t,\pi,b} =  R_{t+1} + \gamma \rho_{t+1}Q_{\pi}(S_{t+1}, A_{t+1}) - Q_{\pi}(S_t, A_t)    
\end{align}
where $\rho_{t+1} = \frac{\pi_\theta(A_t|S_t)}{b(A_t|S_t)}$. On expanding the above expectation:
\begin{equation*}\label{cross_delta_G}
    \begin{split}
    \gamma\sum_{s',a'}P(s'|s,a)b(a'|s')&\E_{b}\Big[\delta_{t,\pi,b} \rho_{t+1}(G_{t+1,\pi,b} \\ &-  Q_\pi(S_{t+1}, A_{t+1}))| S_t=s, A_t=a, S_{t+1}=s', A_{t+1}=a'\Big].   
    \end{split}
\end{equation*}
From \eqref{delta_def}, since $\delta_{t,\pi,b} $ is a function of $(S_t, A_t, S_{t+1}, A_{t+1})$,  $\delta_{t,\pi} $ becomes a constant and can be pulled outside of expectation. Since, $\E_b[\rho_{t+1}(G_{t+1,\pi,b} - Q_\pi(S_{t+1}, A_{t+1}))|S_t=s, A_t=a, S_{t+1}= s', A_{t+1}=a']=0 \,\, \forall \, \{s,s'\} \in S, \{a,a'\} \in A$, therefore the above equation reduces to,
\begin{equation*}
    \begin{split}
    \gamma&\sum_{s',a'}P(s'|s,a)b(a'|s')\delta(s,a,s',a')\times\\
    &\underbrace{\E_{b}\Big[\rho_{t+1}(G_{t+1,\pi,b} -  Q_\pi(S_{t+1}, A_{t+1}))| S_t=s, A_t=a, S_{t+1}=s', A_{t+1}=a'\Big]}_{=0}\\
    &=0    
    \end{split}
\end{equation*}
\end{proof}

\begin{lemma}\label{lemm_delta}
The $\mathbb{E}_b[ \delta_{t,\pi,b} \nabla_\theta \delta_{t,\pi,b} | S_t = s, A_t =a] = 0$.
\end{lemma}
\begin{proof}
On expanding the above expectation,
\begin{equation*}
\begin{split}
& =\sum_{s',a'}P(s'|s,a)b(a'|s') \delta(s,a,s',a') \nabla_\theta \delta(s,a,s',a')\\
& = \sum_{s',a'}P(s'|s,a)b(a'|s')\delta(s,a,s',a')\Big[\nabla_\theta\big(\gamma \rho(s',a')Q_\pi(s',a')\big) - \nabla_\theta Q_\pi(s,a)\Big]\\
& = \sum_{s',a'}P(s'|s,a)b(a'|s')\delta(s,a,s',a')\Big[\nabla_\theta \big(\gamma \rho(s',a')Q_\pi(s',a')\big) - \big( \nabla_\theta r(s,a) + \nabla_\theta(\gamma\rho(s',a')Q_\pi(s',a'))\big)\Big]\\
& = \sum_{s',a'}P(s'|s,a)b(a'|s')\delta(s,a,s',a')\Big[\underbrace{\nabla_\theta \big(\gamma \rho(s',a')Q_\pi(s',a')\big) - \nabla_\theta\big(\gamma\rho(s',a')Q_\pi(s',a')\big)}_{=0}\Big]\\
& = 0
\end{split}
\end{equation*}
\end{proof}

\section{On-policy variance-penalized actor-critic proofs}\label{A:onpac}
Let $\gamma$ be discounting factor. The return $G_{t,\pi}$ under a $\policy$ policy is expressed as:
\begin{align}\label{G_t_onpac}
G_{t,\pi} = R_{t+1} + \gamma G_{t+1,\pi}
\end{align}

\begin{theorem}
\label{thm:bonpac}
(\textbf{On-policy variance in return}) Given a policy $\pi$ and $\bar\gamma = \gamma^2$, the variance in return for a state-action pair $\sigma_\pi(s,a)$ can be computed using Bellman equation, as:
\begin{equation*}
    \begin{split}
    \sigma_\pi(s,a) & = \mathbb{E}_\pi\big[\delta_{t,\pi}^2  + \bar\gamma\sigma_\pi(S_{t+1}, A_{t+1})\big| S_t=s, A_t=a\big]\\
    \text{where, }\delta_{t,\pi} &= R_{t+1} + \gamma Q_\pi(S_{t+1},A_{t+1}) - Q_\pi(S_{t},A_{t}).
    \end{split}
\end{equation*}
\end{theorem}

\begin{proof}
Variance in return is defined as:
\begin{align}\label{sigma_org_onpac}
    \sigma_\pi(s,a) = \mathbb{E}_\pi\Big[\big(G_{t,\pi} - \mathbb{E}_\pi[G_{t,\pi}| S_t=s, A_t =a]\big)^2 \big| S_t=s, A_t =a\Big] \nonumber\\
    \text{where, } \mathbb{E}_\pi[G_{t,\pi}| S_t\!=\!s, A_t\!=\!a] = Q_\pi(s,a).
\end{align}
Using \eqref{G_t_onpac},
\begin{align}\label{G_t_onpac_middle}
    G_{t,\pi} - Q_\pi(S_t,A_t) &=  R_{t+1} + \gamma G_{t+1,\pi} - Q_\pi(S_t,A_t) \nonumber\\
    & = R_{t+1} + \gamma G_{t+1,\pi} - Q_\pi(S_t,A_t) + \gamma Q_\pi(S_{t+1},A_{t+1}) - \gamma Q_\pi(S_{t+1},A_{t+1}) \nonumber\\
    & =  R_{t+1} + \gamma Q_\pi(S_{t+1},A_{t+1}) - Q_\pi(S_t,A_t)  + \gamma\big(G_{t+1,\pi} -  Q_\pi(S_{t+1},A_{t+1})\big) \nonumber\\
    & = \delta_{t,\pi} + \gamma(G_{t+1,\pi} - Q_\pi(S_{t+1}, A_{t+1}))
\end{align}
Substituting \eqref{G_t_onpac_middle} in \eqref{sigma_org_onpac},
\begin{align}\label{sigma_s_a_onpac}
     \sigma_\pi(s,a) & =  \mathbb{E}_\pi\Big[\big(\delta_{t,\pi} + \gamma(G_{t+1,\pi} - Q_\pi(S_{t+1},A_{t+1}))\big)^2|S_t=s, A_t=a\Big] \nonumber\\
     & = \mathbb{E}_\pi\Big[\delta_{t,\pi}^2 + \gamma^2\big(G_{t+1,\pi} - Q_\pi(S_{t+1},A_{t+1})\big)^2 | S_t=s, A_t=a\Big] \nonumber \\
     &\, \, \, \, \, \,  + 2 \mathbb{E}_\pi\Big[\gamma\delta_{t,\pi}\big(G_{t+1,\pi} -  Q_\pi(S_{t+1},A_{t+1})\big)| S_t=s,A_t=a\Big] \nonumber \\
     & = \mathbb{E}_\pi[\delta_{t,\pi}^2 + \gamma^2 \sigma_\pi(S_{t+1},A_{t+1}) | S_t=s,A_t=a] \nonumber\\
     & = \mathbb{E}_\pi[\delta_{t,\pi}^2 + \bar\gamma \sigma_\pi(S_{t+1},A_{t+1}) | S_t=s,A_t=a]
\end{align}
The above equation is justified using Lemma \ref{off_policy_cross_term}, where let $\rho_{t+1} =1$ (importance sampling ratio) as the target policy is equivalent to the behavior policy $\pi =b$.
\end{proof}

\begin{theorem}
\label{thm:Jonpac}
(\textbf{Variance-penalized on-policy PG theorem}): Given $s_{0} \sim d_{0}$ - an initial state distribution,  a stochastic policy $\pi_\theta$, $\bar\gamma = \gamma^2$, the gradient of on-policy mean-variance objective $J$ w.r.t. $\theta$ is given by
\begin{align*}
    \mathbb{E}_{s_0 \sim d_0}\Bigg[\sum_{k=0}^{\infty}\sum_{s}\Big[& P_{\gamma}^{(k)}(s|s_0)\sum_{a}\nabla_\theta \pi_\theta(a|s) Q_\pi(s,a)
      - \psi P_{\bar\gamma}^{(k)}(s|s_0)\sum_{a}\nabla_\theta \pi_\theta(a|s) \sigma_\pi(s,a)\Big]\Bigg].
\end{align*}
\end{theorem}

\begin{proof}
Let $\sigma_\pi(s) = \sum_{a} \pi_\theta(a|s) \sigma_\pi(s,a)$. We first solve for the gradient of variance in the return $\sigma_\pi(s)$ w.r.t. $\theta$ using \eqref{sigma_s_a_onpac}
\begin{align}\label{sig_grad_opac}
\nabla_\theta \sigma_\pi(s) &= \sum_{a} \nabla_\theta \pi_\theta(a|s) \sigma_\pi(s,a) +\sum_{a} \pi_\theta(a|s) \nabla_\theta \sigma_\pi(s,a) \nonumber \\
&=\sum_{a} \nabla_\theta \pi_\theta(a|s) \sigma_\pi(s,a)   \nonumber \\
& \, \, \, \, \, \, \, \, \, \,+\sum_{a} \pi_\theta(a|s)\Big[ \mathbb{E}_\pi\big[2\delta_{t,\pi} \nabla_\theta \delta_{t,\pi}| S_t =s, A_t=a\big] + \bar{\gamma} \sum_{s'}P(s'|s,a) \nabla_\theta \sigma_\pi(s')\Big] \nonumber\\
& = \sum_{a} \nabla_\theta \pi_\theta(a|s) \sigma_\pi(s,a) + \bar{\gamma} \sum_{a} \pi_\theta(a|s) \sum_{s'}P(s'|s,a) \nabla_\theta \sigma_\pi(s')
\end{align}
where $\mathbb{E}_\pi[2\delta_{t,\pi} \nabla_\theta \delta_{t,\pi}| S_t\! =\!s, A_t\!=\!a] =0$ following  Lemma \ref{lemm_delta} setting $\rho =1$ and $b = \pi$. 
Let one-step transition from $s$ to $s'$ be denoted by,
\begin{align}\label{discounted_P_onP}
 P_{\bar{\gamma}}^{(1)}(s'|s) = \bar{\gamma} \sum_{a} \pi_\theta(a|s) P(s'|s,a) .  
\end{align}
Using \eqref{discounted_P_onP} and unrolling $\nabla_\theta \sigma_\pi(s')$ using \eqref{sig_grad_opac},
\begin{align}\label{grad_sigma_onpac}
\nabla_\theta \sigma_{\pi}(s) &=\sum_{k=0}^{\infty}\sum_{s'}P_{\bar{\gamma}}^{(k)}(s'|s) \sum_{a'} \nabla_\theta \pi_\theta(a'|s') \sigma_\pi(s',a').
\end{align}
where $P_{\bar{\gamma}}^{(k)}(s'|s)$ follows Definition 2 in the paper. Similarly, the value function is $V_\pi(s) = \sum_{a}\pi_\theta(a|s) Q_\pi(s,a)$. Following the policy gradient theorem, the gradient of $V_\pi(s)$ w.r.t. $\theta$,
\begin{align}\label{grad_V_onpac}
\nabla_\theta V_\pi(s) = \sum_{k=0}^{\infty}\sum_{s'}P_{\gamma}^{(k)}(s'|s) \sum_{a'} \nabla_\theta \pi_\theta(a'|s') Q_\pi(s',a').
\end{align}
Assuming start state is $s_0 \sim d_0$, combining the equations \eqref{grad_sigma_onpac} and \eqref{grad_V_onpac}, the gradient of objective $J$ in (3) becomes:
\begin{align*}
       \nabla_{\theta} J_{d_0}(\theta) = \mathbb{E}_{s_0 \sim d_0}\Bigg[\sum_{k=0}^{\infty}\sum_{s}\Big[& P_{\gamma}^{(k)}(s|s_0)\sum_{a}\nabla_\theta \pi_\theta(a|s) Q_\pi(s,a)\\
      &- \psi P_{\bar\gamma}^{(k)}(s|s_0)\sum_{a}\nabla_\theta \pi_\theta(a|s) \sigma_\pi(s,a)\Big]\Bigg].
\end{align*}
\end{proof}

\section{Off-policy variance-penalized actor-critic proofs}\label{A:OffPAC}
The return $G_{t,\pi,b}$ is the estimation of state-action value under a target policy $\pi$ following a behavior $b$ policy (with correction factor enrolled inside).
\begin{align}\label{eq:off_pol_return}
G_{t,\pi,b} = R_{t+1} +  \gamma \rho_{t+1} G_{t+1,\pi,b}
\end{align}
The correction factor $\rho$ does not appear with the reward term because the expression for return $G_{t,\pi,b}$ is expanded given a state $S_t$ and action $A_t$. As, $A_t$ is given and not sampled from the behavior policy, the correction factor does not appear before the reward term.

\begin{theorem}
\label{thm:sigma_th_offpac}
(\textbf{Off-policy variance in the return})
Given a behaviour policy $b$ and $\bar\gamma = \gamma^2$, the off-policy variance in return for a state-action pair $\sigma_\pi(s,a)$ can be computed using Bellman equation as follows:
\begin{align}
\sigma_\pi(s,a) &= \mathbb{E}_b\big[\delta_{t,\pi}^2\!+\! {\bar\gamma}\rho_{t+1}^2 \sigma_\pi(S_{t+1}, A_{t+1})\big| S_t\!=\!s,A_t\!=\!a\big],\label{variance_offp}\\
\text{where, }\delta_{t,\pi} &= R_{t+1} + \gamma \rho_{t+1} Q_{\pi}(S_{t+1}, A_{t+1}) - Q_\pi(S_{t}, A_{t}). \label{var_td_off_policy}
\end{align}
\end{theorem}

\begin{proof}
Variance in the return is:
\begin{align}\label{3}
\sigma_\pi(s,a) & = \mathbb{E}_{b}
\Big[\big(G_{t,\pi,b} - \mathbb{E}_{b}[G_{t,\pi,b}|S_t = s, A_t=a]\big)^{2} \Big| S_t = s, A_t=a\Big],
\end{align}
where, $\mathbb{E}_{b}[G_{t,\pi,b} | S_t\! =\!s, A_t\!=\!a ] = Q_\pi(s, a)$ (under $\pi$ policy because correction factor is rolled inside the $G_{t,\pi,b}$). Let : \[\delta_{t,\pi,b} = R_{t+1} + \gamma \rho_{t+1} Q_{\pi}(S_{t+1}, A_{t+1}) - Q_\pi(S_{t}, A_{t}).\]
Using \eqref{eq:off_pol_return},
\begin{equation}\label{var_middle}
\begin{split}
    G_{t,\pi,b} &- Q_\pi(S_t, A_t) \\
    =&  R_{t+1} +  \gamma \rho_{t+1} G_{t+1,\pi,b} - Q_\pi(S_t, A_t) + \gamma \rho_{t+1} Q_{\pi}(S_{t+1}, A_{t+1}) - \gamma \rho_{t+1} Q_{\pi}(S_{t+1}, A_{t+1})\\
    =&  \underbrace{R_{t+1} +  \gamma \rho_{t+1} Q_{\pi}(S_{t+1}, A_{t+1}) - Q_\pi(S_t, A_t)}_{\delta_{t,\pi,b}} +  \gamma \rho_{t+1} (G_{t+1,\pi,b} -  Q_{\pi}(S_{t+1}, A_{t+1}))\\
    =& \delta_{t,\pi,b} + \gamma\rho_{t+1}\Big(G_{t+1,\pi,b} -  Q_{\pi}(S_{t+1},A_{t+1}) \Big)
\end{split}
\end{equation}
Substituting \eqref{var_middle} in \eqref{3},
\begin{equation}\label{sigma_med}
\begin{split}
    \sigma_\pi(s,a) =& \mathbb{E}_b\Bigg[\Big(\delta_{t,\pi,b} + \gamma\rho_{t+1}( G_{t+1, \pi,b} - Q_{\pi}(S_{t+1},A_{t+1}))\Big)^{2} \Bigg| S_t\!=s, A_t\!=a\Bigg]\\
    =& \mathbb{E}_b\big[\delta_{t,\pi,b}^2\! +\! \gamma ^2   \rho_{t+1}^2 \big( G_{t+1,\pi,b}\!- Q_{\pi}(S_{t+1},A_{t+1})\big)^{2} \big| S_t\! = s, A_t\!=a\big] \\
    & + 2 \mathbb{E}_b\big[\gamma\delta_{t,\pi,b}\rho_{t+1} ( G_{t+1,\pi,b} -  Q_{\pi}(S_{t+1},A_{t+1})) \big| S_t\! = s, A_t\!=a\big].
\end{split}
\end{equation}
Using Lemma \ref{off_policy_cross_term}, we conclude that the last term in \eqref{sigma_med} is equivalent to zero. Therefore, the above term reduces as follows:
\begin{align*}
    \sigma_\pi(s,a) = \mathbb{E}_b\big[\delta_{t,\pi,b}^2 + \gamma ^2 \rho_{t+1}^2 \sigma_\pi(S_{t+1}, A_{t+1}) \big| S_t = s, A_t=a\big].
\end{align*}
\end{proof}

\begin{theorem}
\label{thm:Jofpac}
(\textbf{Variance-penalized off-policy PG theorem}): Given $s_0 \sim d_0$, $a_0 \sim b$, a stochastic target policy $\pi_\theta$, $T_\gamma^{(k)}(s,a|s_0,a_0)$ and $\bar{T}_{\bar\gamma}^{(k)}(s,a|s_0,a_0)$ following Definition 3 and 4 respectively, the gradient of off-policy mean-variance objective function $J$ w.r.t. $\theta$ is 
\begin{equation*}
\begin{split}
    \nabla_{\theta}J_{d_{0}}(\theta) = \mathbbm{E}_{s_0 \sim d_0, a_0 \sim b}\Bigg[\sum_{k=0}^{\infty} \sum_{s,a}\! \Big[&T_\gamma^{(k)}(s,a|s_0,a_0) \nabla_\theta\log\pi_\theta(a|s)Q_\pi(s,a)\\
    &- \psi \bar{T}_{\bar\gamma}^{(k)}(s,a|s_0,a_0)
        [1 + \mathbbm{1}_{k\geq 1}]\nabla_\theta \log\pi_\theta(a|s)\sigma_\pi(s,a)\Big]\Bigg].
\end{split}
\end{equation*}
\end{theorem}

\begin{proof}
We define the objective function J, for the off-policy variance penalized actor-critic as:
\begin{equation}\label{objective_offpac}
\begin{split}
    J_{d_0}(\theta) & = \mathbb{E}_{s \sim d_0}\Bigg[\sum_{a} \pi_{\theta}(a|s)\big[ Q_\pi(s, a) - \psi \sigma_\pi(s,a)\big] \Bigg]\\
    & = \mathbb{E}_{s \sim d_0, a \sim b} \Bigg[\rho(s,a)\big[Q_\pi(s, a) - \psi \sigma_\pi(s,a))\big] \Bigg].
\end{split}
\end{equation}
where $d_0$ is the initial state distribution. The expectation over actions drawn from the target policy is converted to that over the behavior policy $b$ by considering a correction factor inside the expectation term. The gradient of $\rho(s,a)$ is equivalent to:
\begin{equation}\label{grad_rho}
\begin{split}
    \nabla_{\tht} \rho(s,a) &=  \frac{\nabla_{\tht} \pi_\theta(a|s)}{b(a|s)}\\
    & = \frac{\pi_\theta(a|s)}{b(a|s)} \times \frac{\nabla_{\tht} \pi_\theta(a|s)}{\pi_\theta(a|s)} \\
    & = \rho(s,a)\nabla_{\tht} \log\pi_\theta(a|s).
\end{split}
\end{equation}
The gradient of $J$ w.r.t. $\theta$ following \eqref{objective_offpac} and \eqref{grad_rho} is:
\begin{align}\label{J_off_pol_3}
\nabla_\theta J_{d_0}(\theta) & = \mathbb{E}_{{s \sim d_0},a\sim b} \Big[\nabla_\theta \rho(s,a) \big(Q_\pi(s,a) - \psi \sigma_\pi(s,a) \big) + \rho(s,a) \big(\nabla_\theta Q_\pi(s,a) - \psi \nabla_\theta \sigma_\pi(s,a) \big)\Big]\nonumber\\
&= \mathbb{E}_{{s \sim d_0},a\sim b} \big[ \rho(s,a)\nabla_\theta \log\pi_\theta(a|s) Q_\pi(s,a) + \rho(s,a)\nabla_\theta Q_\pi(s,a) \big]\nonumber\\
&\quad -\psi\mathbb{E}_{{s \sim d_0},a\sim b} \big[ \rho(s,a)\nabla_\theta \log\pi_\theta(a|s) \sigma_\pi(s,a) + 
\rho(s,a)\nabla_\theta\sigma_\pi(s,a) \big].
\end{align}
Here, we first solve for the gradient of variance $\sigma_\pi(s,a)$ with respect to policy parameter $\theta$. Let $\bar\gamma = \gamma^2$. Using Theorem \ref{thm:sigma_th_offpac},
\begin{equation}\label{_grad_sigma_middle}
\begin{split}
    \sigma_\pi(s,a) =& \mathbb{E}_b[\delta_{t,\pi,b}^2|S_t=s, A_t=a] + \bar{\gamma}\sum_{s'}P(s'|s,a) \sum_{a'}b(a'|s') \rho(s',a')^2 \sigma_\pi(s',a')\\
    \nabla_\theta \sigma_\pi(s,a) =& \,\,\underbrace{2\mathbb{E}_b[ \delta_{t,\pi,b} \nabla_\theta \delta_{t,\pi,b} | S_t=s, A_t\!=\!a]}_{=0} \\
    & + \bar\gamma \sum_{s'}P(s'|s,a)\sum_{a'}b(a'|s') \big\{2 \rho(s',a')\nabla_\theta \rho(s',a') \sigma_{\pi}(s',a') +\rho(s',a')^2 \nabla_\theta \sigma_{\pi}(s',a')\big\} \\
    =& \bar\gamma \sum_{s'}P(s'|s,a) \sum_{a'}b(a'|s') \rho(s',a')^2 \big\{2 \frac{\nabla_\theta \rho(s',a')}{\rho(s',a')}  \sigma(s',a') \\
    &\quad\quad +  \nabla_\theta \sigma(s',a')\big\} \quad{\text{[Using Lemma \ref{lemm_delta}]}}\\
    =& \bar\gamma \sum_{s'}P(s'|s,a) \sum_{a'}b(a'|s') \rho(s',a')^2 \big\{2 \nabla_\theta \log(\pi_\theta(a'|s')) \sigma(s',a') +  \nabla_\theta \sigma(s',a')\big\}
\end{split}
\end{equation}
Using Lemma \ref{lemm_delta}, $2\mathbb{E}_b[ \delta_{t,\pi} \nabla_\theta \delta_{t,\pi} | S_t = s, A_t =a] = 0$. Let:\\
\begin{align}\label{P_sigma}
    \bar{T}_{\bar\gamma}^{(1)}(S_{t+1},A_{t+1}| S_t,A_t) = \bar\gamma {P}(S_{t+1}|S_t,A_t) b(A_{t+1}|S_{t+1}) \rho_{t+1}^2.
\end{align}
Similarly for k-steps, the above term can be represented as,  
\begin{align}\label{kstep_P}
 \bar{T}_{\bar\gamma}^{(k)}(S_{t+k},A_{t+k}| S_t,A_t) = \bar{T}_{\bar\gamma}^{(1)}(S_{t+k},A_{t+k}| S_{t+k-1},A_{t+k-1}) \bar{T}_{\bar\gamma}^{(k-1)}(S_{t+k-1},A_{t+k-1}| S_{t},A_{t}).   
\end{align}
Using \eqref{_grad_sigma_middle}, \eqref{P_sigma} and \eqref{kstep_P}, the gradient of $\sigma_\pi(s,a)$ can be further simplified as:
\begin{align}
\nabla_\theta \sigma_\pi(s,a) =& \sum_{s',a'} \bar{T}_{\bar\gamma}^{(1)}(s',a'| s,a) \{2 \nabla_\theta \log(\pi_\theta(a'|s')) \sigma_\pi(s',a') +  \nabla_\theta \sigma_\pi(s',a')\} \nonumber\\
=& 2 \sum_{k=1}^{\infty}\sum_{s',a'} \bar{T}_{\bar\gamma}^{(k)}(s',a'| s,a) \{ \nabla_\theta \log(\pi_\theta(a'|s')) \sigma_\pi(s',a')\}.
\end{align}

Let $\bar{T}_{\bar \gamma}^{(0)}(s,a|s,a) =\rho(s,a)$. Solving for the second term in the gradient of $J_{d_0}(\theta)$ in \eqref{J_off_pol_3},
\begin{align}\label{grad_sigma_offP_new}
\rho(s,a)\sigma_\pi(s,a)\nabla_\theta& \log(\pi_\theta(a|s)) + 
\rho(s,a)\nabla_\theta\sigma_\pi(s,a) \nonumber\\
= & \bar{T}_{\bar\gamma}^{(k=0)}(s,a|s,a)\Big[\nabla_\theta \log(\pi_\theta(a|s))\sigma_\pi(s,a) + 
\nabla_\theta\sigma_\pi(s,a)\Big] \nonumber\\
= &  \sum_{k=0}^{\infty}\sum_{s',a'}\bar{T}_{\bar\gamma}^{(k)}(s',a'|s,a)[1 + \mathbbm{1}_{k\geq 1}]\Big[\nabla_\theta \log(\pi_\theta(a'|s'))\sigma_\pi(s',a')\Big]
\end{align}
Now, we are solving for the gradient of $Q_\pi(s,a)$ w.r.t. $\theta$ parameter of the policy in off-policy fashion.
\begin{align}\label{grad_Q_off_pol}
Q_\pi(s,a) =& r(s,a) + \gamma \sum_{s'}P(s'|s,a) \sum_{a'}b(a'|s') \rho(s',a') Q_\pi(s',a') \nonumber\\
\nabla_\theta Q_\pi(s,a) =& \gamma \sum_{s'}P(s'|s,a) \sum_{a'}b(a'|s') \Big\{\nabla_\theta\rho(s',a') Q_\pi(s',a') + \rho(s',a') \nabla_\theta Q_\pi(s',a')\Big\} \nonumber\\
= & \gamma \sum_{s'}P(s'|s,a) \sum_{a'}b(a'|s')\rho(s',a') \Big\{ \nabla_\theta \log(\pi_\theta(a'|s'))Q_\pi(s',a') + \nabla_\theta Q_\pi(s',a')\Big\}
\end{align}
Let:
\begin{align}\label{P_Q}
 T_{\gamma}^{(1)}(S_{t+1},A_{t+1}|S_{t},A_{t}) = \gamma P(S_{t+1}|S_{t},A_t) b(A_{t+1}|S_{t+1}) \rho_{t+1}.   
\end{align}
Using the above equation in \eqref{grad_Q_off_pol}, the gradient of $Q_\pi(s,a)$ is further simplified as:
\begin{align}\label{grad_Q_offpac}
\nabla_\theta Q_\pi(s,a) =&  \sum_{s',a'}T_{\gamma}^{(1)}(s',a'|s,a) \{ \nabla_\theta \log(\pi_\theta(a'|s'))Q_\pi(s',a') + \nabla_\theta Q_\pi(s',a')\} \nonumber\\
=& \sum_{k=1}^{\infty}\sum_{s',a'} T_{\gamma}^{(k)}(s',a'|s,a) \{ \nabla_\theta \log(\pi_\theta(a'|s'))Q_\pi(s',a')\}
\end{align}
Now taking the first term in gradient of $J_d(\theta)$ in \eqref{J_off_pol_3},
\begin{align}\label{grad_Q_offP}
\rho(s,a) Q_\pi(s,a)\nabla_\theta \log(\pi_\theta(a|s)) & + \rho(s,a)\nabla_\theta Q_\pi(s,a) \nonumber\\
=&   \sum_{s',a'} T_\gamma^{(k=0)}(s',a'|s,a) \{\nabla_\theta \log(\pi_\theta(a|s))Q_\pi(s,a) + \nabla_\theta Q_\pi(s,a)\} \nonumber\\
= &\sum_{k=0}^{\infty} \sum_{s',a'} T_\gamma^{(k)}(s',a'|s,a) \{\nabla_\theta \log(\pi_\theta(a'|s'))Q_\pi(s',a')\}
\end{align}
Let initial state $s_0 \sim d_0$ and initial action be $a_0 \sim b$, then the gradient of $J_{d_0}(\theta)$ w.r.t. $\theta$ in \eqref{J_off_pol_3} can be written using \eqref{grad_sigma_offP_new} and \eqref{grad_Q_offP} as
\begin{equation*}
\begin{split}
    \nabla_{\theta}J_{d_{0}}(\theta) = \mathbbm{E}_{s_0 \sim d_0, a_0 \sim b}\Bigg[\sum_{k=0}^{\infty} \sum_{s,a}\! \Big[&T_\gamma^{(k)}(s,a|s_0,a_0) \nabla_\theta\log\pi_\theta(a|s)Q_\pi(s,a)\\
    &- \psi \bar{T}_{\bar\gamma}^{(k)}(s,a|s_0,a_0)
        [1 + \mathbbm{1}_{k\geq 1}]\nabla_\theta \log\pi_\theta(a|s)\sigma_\pi(s,a)\Big]\Bigg].
\end{split}
\end{equation*}
Here,
\begin{align*}
T_{\gamma}^{(0)}(s,a|s,a) =& \rho(s,a),\\
\bar{T}_{\bar\gamma}^{(0)}(s,a| s,a) =& \rho(s,a),\\
T_{\gamma}^{(1)}(s',a'|s,a) =& T_{\gamma}^{(0)}(s,a|s,a)[\gamma P(s'|s,a) b(a'|s') \rho(s',a')],    \\
\bar{T}_{\bar\gamma}^{(1)}(s',a'| s,a) =& \bar{T}_{\bar\gamma}^{(0)}(s,a| s,a) [\bar\gamma P(s'|s,a) b(a'|s') \rho(s',a')^2].
\end{align*}
\end{proof}


\section{Convergence analysis}\label{app:convergence_analysis}
 
We will first present few definitions here that would be used later in the proofs. Following the notations in \cite{Bertsekas1996NP}, let $r \in \R^n$ be a vector whose $i^{th}$ component is denoted by $r(i)$. 
\begin{definition}
Given a vector $r$, for a positive vector $\xi = (\xi(1), \xi(2), \dots, \xi(n))$, the \textit{weighted maximum norm} $\norm{.}_{\xi}$ is denoted by \[\norm{r}_{\xi} = \max_i \frac{r(i)}{\xi(i)}.\]
\end{definition}
\begin{definition}
The function $H: \R^n \xrightarrow{} \R^n$ is a \textit{weighted maximum norm pseudo-contraction} if there exist $r^* \in \R^n$ and a constant $\beta \in [0,1)$ such that
\begin{align}\label{H_contraction}
\norm{Hr - r^*}_\xi \leq \beta \norm{r - r^*}_\xi.
\end{align}
Here $r^* = Hr^*$ is the fixed point of H.
\end{definition}
\begin{definition}
Let $\{\F_n\}_{n \geq 0}$ be the increasing sequence of $\sigma$-algebra (called \textit{filtration}) such that 
\[\F_0 \subset \F_1 \subset \F_2 \dots \subset \F_n.\]
\end{definition}
\begin{definition}
$X_0, X_1, \dots$ is said to be \textit{martingale} relative to filtration $\{\F_n\}_{n \geq 0}$ if for every $n$ it follows:
\[\E[X_{\timestep + 1} \mid \F_n] = X_{\timestep}.\]
\end{definition}{}
\begin{definition}\label{def:martingale_difference}
$\{\martingalediff_n\}_{n\geq 0}$, such that $\martingalediff_{\timestep+1} = X_{\timestep+1} - X_{\timestep}$ is called the \textit{martingale difference} sequence with $\martingalediff_{0} = X_{0}$. It follows:
\[\E[\martingalediff_{\timestep+1} \mid \F_n] = 0.\]
\end{definition}

\begin{definition}\label{def:proper_policy}
A stationary policy $\pi$ is said to be a \textit{proper policy}, if there is a positive probability of reaching a termination state after atmost n steps irrespective of the initial start state.
\end{definition}

\subsection{Stochastic Approximation}\label{section:SA}
In this section, we would briefly provide pre-requisite on Stochastic Approximation that is needed to understand our work on convergence analysis later.

In order to solve an optimization problem using iterative algorithms, many times all the information required to solve it is not available. One sometimes has to work with noisy version of the available data (for example, working with an estimate of an expectation). Stochastic Iterative Algorithms can work in presence of noisy iterates. Lets say, we want to solve a system of equation of the form,
\[ Hr= r,\quad  r \in \R^n,\] where $H: \R^n \xrightarrow{} \R^n$ is a mapping to itself. $r^*$ is a \textit{fixed point} of $H$ and satisfies $Hr^* = r^*$. Let $(Hr)(i)$ denote the $i^{th}$ component of $Hr \, \forall r \in \R^n$. The stochastic approximation algorithm in simplest form is given by:
\begin{align}\label{noisy_sa}
    r_{t+1}(i) = (1-\gamma_t(s))r_t(i) + \gamma_t(s)\big((H_t r_t)(i) + w_t(i) + u_t(i)\big)
\end{align}
where $t$ denotes the time step i.e. $t \in \{0,1,\dots ,n\}$, $\gamma_t(s)$ the step size, and $w_t(i)$ is the random noise with zero mean and $u_t(i)$ is additional noise with not necessarily zero mean. Here $H_t$ denotes the iteration mapping that is allowed to change from iteration to another. Let $\F$ be the history of algorithm until $t$ time i.e., \[\F_t = \{r_0(i), \dots, r_t(i), w_0(i), \dots ,w_{t-1}(i), u_0(i), \dots ,u_{t-1}(i), \gamma_0(i), \dots, \gamma_t(s)\}.\] We assume that $\F_t$ is an increasing sequence such that $\F_{t+1}$ always contain history $\F_t$.

There are many different tools to analyse the convergence results of a stochastic iterations, but here we will reiterate Proposition 4.5 from \cite{Bertsekas1996NP} to show the convergence of the iteration \eqref{noisy_sa} to a fixed point.

\begin{proposition}\label{prop:basic_sa_convergence}
If $r_t$ is the sequence generated from \eqref{noisy_sa}. If we assume the following statements:
\begin{enumerate}
    \item The stepsize $\gamma_t(i)$ are non-negative and satisfy: \[\sum_{t=0}^{\infty} \gamma_t(i) = \infty, \quad \sum_{t=0}^{\infty} \gamma_t(i)^2 < \infty.\]
    \item The noisy term $w_t(i)$ follows $\E[w_t(i)|\F_t] =0$ and $\E[w_t(i)^2 | \F_t] \leq A+ B \norm{r_t}^2$ where A and B are some constants.
    \item $H$ is a weighted maximum norm pseudo-contraction mapping \eqref{H_contraction} \[\norm{H_t r_t - r^*}_\xi \leq \beta \norm{r_t - r^*}_\xi.\]
    \item There exist non-negative random sequence $\theta_t$ that converges to zero with probability (w.p.) 1 and follows  for all t, \[|u_t(i)| \leq \theta_t (\norm{r_t}_\xi + 1).\]
\end{enumerate}
Then $r_t$ converges to $r^*$ w.p. 1.
\end{proposition}

\subsection{Multiple timescale analysis}
Similar to the two timescales analysis by \cite{borkar1997stochastic}, we will do a three timescales analysis on the value function, the variance function and the policy to show the convergence of the proposed algorithm. Here we will write the generic analysis with the three timescales and then later will analyse how it fits our proposed method. 

We are interested in the  ordinary differential equation (ODE) approach to analyse three stochastic approximations:

\begin{align}
    x_{n+1} &= x_{n} + a(n)[L_{1}(x_{n}, z_{n}) + M_{n+1}^{(1)}],\label{x}\\
    y_{n+1} &= y_{n} + b(n)[L_{2}(x_{n}, y_{n}, z_{n}) + M_{n+1}^{(2)}],\label{y}\\
    z_{n+1} &= z_{n} + c(n)[L_{3}(x_{n}, y_{n}, z_{n}) + M_{n+1}^{(3)}],\label{z}
\end{align}
where, $n \geq 0$, $x \in \R^d$, $y \in \R^k$, $z \in \R^m$, $L_{i}, \, \forall i \in \{1,2,3\}$ are Lipschitz functions, ${M_{n}^{(i)}},\, \forall i \in \{1,2,3\}$ are Martingale Difference Sequences with respect to increasing $\sigma$-field \[\F_n=\sigma(x_i, y_i, z_i, M_{i}^{(1)}, M_{i}^{(2)}, M_{i}^{(3)}, i\leq n).\] The step-sizes $\{a(n)\}, \{b(n)\}, \{c(n)\}$ are positive scalars satisfying:
\begin{align}\label{step-size law}
  \sum_{n}a(n)=\infty, \sum_{n}b(n)=\infty, \sum_{n}c(n)=\infty, \sum_{n}[a(n)^2+ b(n)^2+ c(n)^2]<\infty.  
\end{align}
Further the relationships between the step sizes satisfy:
\begin{equation}\label{stepsize_three_level}
  \frac{c(n)}{b(n)} \xrightarrow[]{} 0, \frac{b(n)}{a(n)} \xrightarrow[]{} 0.  
\end{equation}
The above condition imposes that \eqref{x} is moving at the fastest timescale,  followed by \eqref{y} and then \eqref{z} is moving at the slowest time scale. Thus $x(\cdot)$ is the fastest component as compared to $y(\cdot), z(\cdot)$. Therefore, one can interpret $y(\cdot)$ and $z(\cdot)$ as \emph{quasi-static} (like constant) when analysing the equation for $x(\cdot)$. Therefore, one can see the o.d.e. of $x(\cdot)$ as
\begin{align}\label{x_ode}
    \dot x(t) = L_{1}(x(t), z) 
\end{align}
where $z$ is a constant (due to quasi-static behavior of $z(.)$).

\begin{assumptionF}\label{assump:MTA_1}
$\sup_{n}\, [\norm{x_n}{} + \norm{y_n}{} + \norm{z_n}{}] < \infty, \text{ w.p. } 1.$
\end{assumptionF}
The iterates of \eqref{x}, \eqref{y} and \eqref{z} are bounded with probability 1.

\begin{assumptionF}\label{assump:MTA_2}
The o.d.e. of $x(\cdot)$ in \eqref{x_ode} has a globally asymptotically stable equilibrium $\lambda_{1}(z)$ where $\lambda_{1}: \R^m \xrightarrow{} \R^d$ is a Lipschitz function.
\end{assumptionF}
$x(t)$ in \eqref{x} tracks asymptotically $\lambda_{1}(z(t'))$ for $t>0$ and here $t'$ is moving at a very slow timescale as compared to $t$. As $y(\cdot)$ is moving at a faster timescale than $z(\cdot)$, therefore, $y(\cdot)$ observes $z(\cdot)$ as quasi-static. Similarly, $y(.)$ is moving at a slower timescale as compared to $x(\cdot)$, thus, $y(\cdot)$ sees $x(\cdot)$ as nearly \emph{equilibrated}. Now looking at the o.d.e. for $y(\cdot)$ in \eqref{y}, 
\begin{align}\label{y_ode}
    \dot y(t) = L_{2}(\lambda_{1}(z), y(t), z)
\end{align}
where $z$ is held as a constant (quasi-static) and $x$ as equilibrated $\lambda_{1}(z)$.

\begin{assumptionF}\label{assump:MTA_3}
The o.d.e. of $y(\cdot)$ in \eqref{y_ode} has a globally asymptotically stable equilibrium $\lambda_{2}(\lambda_{1}(z),z)$ where $\lambda_{2}: \R^{d+m} \xrightarrow{} \R^{k}$ is a Lipschitz function.
\end{assumptionF}
$y(t)$ in \eqref{y} tracks asymptotically $\lambda_{2}(\lambda_{1}(z(t')), z(t'))$ for $t>0$ and here $t'$ is moving at a very slow timescale as compared to $t$.  Now, lets look at the o.d.e. of $z(t)$:
\begin{align}\label{z_ode}
    \dot z(t) = L_{3}\Big(\lambda_{1}\big(z(t)\big),\, \lambda_{2}\big(\lambda_{1}(z(t)),z(t)\big),\, z(t)\Big),
\end{align}
where both $x$ and $y$ have equilibrated to $\lambda_{1}(z(t))$ and $ \lambda_{2}(\lambda_{1}(z(t)),z(t))$ respectively (due to difference in the timescales).

\begin{assumptionF}\label{assump:MTA_4}
The o.d.e. of $z(\cdot)$ in \eqref{z_ode} has a globally asymptotically stable equilibrium $z^{*}$.
\end{assumptionF}

Using \textbf{Lemma 1} in (\cite{borkar2009stochastic}, Pg 66, Ch 6), we know that $\norm{x_{n} - \lambda_1(z_n)}{} \xrightarrow{} 0$ w.p. 1, or one could say that $\{x_n\}$ sequence asymptotically tracks ${\lambda_{1}(z_n)}$ almost surely. With the similar arguments of Lemma 1 (as before), we can say also that $\norm{y_{n} - \lambda_2(\lambda_1(z_n), z_n)}{} \xrightarrow{} 0$ w.p. 1. Furthermore, using \textbf{Theorem 2} in (\cite{borkar2009stochastic}, Pg 66, Ch 6), we can easily see that $(x_n, y_n, z_n) \xrightarrow{} (\lambda_1(z^{*}), \, \lambda_2(\lambda_1(z^{*}), z^{*}), \, z^{*})$. This completes the proof for the multiple timescale analysis.

\subsection{Actor-Critic convergence using multiple timescale analysis}
We use \textit{Softmax} policy parameterization, where for some $\theta \in \R^{|S| \times |A|}$,
\begin{align*}
    \poltheta(a|s) = \frac{\exp(\theta_{s,a})}{\sum_{a' \in A}\exp(\theta_{s,a'})} \quad \forall s \in S, a \in A.
\end{align*}

\begin{equation} \label{grad_softmax}
    \begin{split}
        \text{ Let, } f_{s,a}(x) &= \frac{\exp(\theta_{s,a})}{\sum_{a' \in A}\exp(\theta_{s,a'})} = \frac{p(x)}{q(x)}\\
        \frac{\partial f_{s,a}(x)}{\partial \theta_{s', a'}} &= \frac{p'(x)q(x) - p(x)q'(x)}{q(x)^2}\\
        \frac{\partial f_{s,a}(x)}{\partial \theta_{s', a'}} &=
        \begin{cases}
            f_{s,a}(x) (1 - f_{s,a'}(x)),& \text{if } s=s', a = a',\\
            -f_{s,a}(x)f_{s,a'}(x),              & \text{if } s=s', a \neq a',\\
            0,                                   & \text{if } s\neq s'.
        \end{cases}
    \end{split}
\end{equation}

Let $S$ and $A = \cup_{s \in S} A(s)$ be finite set of states and corresponding actions. Let $Q_{n}(.,.), \variance_{n}(.,.)$ denote the $n^{th}$ update of the state-action value function and variance function respectively. Let $\{Y_n(s,a)\}$ be i.i.d. random variable with distribution as $p(.|s, a)$ (transition). Let $\actionsample_{n}(i)$ denote the action chosen following policy $\poltheta^{n}(i,.)$. Let $\{a(n)\}, \{b(n)\},\{c(n)\}$ be diminishing step sizes satisfying \eqref{step-size law}. The TD error is given by, \[ \delta_n(s,a, Y_n(s, a),\actionsample_{n}(Y_{n}(s,a))) = r(s,a) + Q_{n}(Y_{n}(s,a), \actionsample_{n}(Y_{n}(s,a))) - Q_{n}(s,a).\] The \textbf{actor-critic algorithm} is as follows:

\begin{align}
    Q_{n+1}(s,a) &=  Q_{n}(s,a) + a(n)\Big(r(s,a) + Q_{n}\big(Y_{n}(s,a), \actionsample_{n}(Y_{n}(s,a))\big) - Q_{n}(s,a)\Big), \label{Q_stochastic}\\
    \variance_{n+1}(s,a) &= \variance_{n}(s,a) + b(n) \Big( \delta_n\big(s,a, Y_n(s,a),\actionsample_{n}(Y_{n}(s,a))\big)^2 \nonumber\\
    &\quad + \variance_{n}\big(Y_n(s,a),\actionsample_{n}(Y_{n}(s,a))\big) - \variance_{n}(s,a) \Big), \label{variance_stochastic}\\
    \theta_{n+1} &= \theta_{n} + c(n)\Big(\nabla_{\theta} \log \poltheta^{n}(a|s) (Q_{n}(s,a) - \psi \variance_{n}(s,a))\Big), \quad \text{where,}  \psi \in \R. \label{policy_stochastic}
\end{align}
First note that using \eqref{grad_softmax}:
\begin{equation}\label{grad_log_pi}
    \begin{split}
        \frac{\partial \log \poltheta(a|s)}{\partial \theta_{s',a'}} &= \identity[s=s'] \bigg\{\frac{1}{\poltheta(a|s)} \times \frac{\partial \poltheta(a|s)}{\partial \theta_{s,a'}}\bigg\},\\
        &= \identity[s=s'] \bigg\{\frac{1}{\poltheta(a|s)} \times \poltheta(a|s) [\identity[a=a'] - \poltheta(a'|s)]\bigg\},\\
        &= \identity[s=s'](\identity[a=a'] - \poltheta(a'|s)), \quad s, s' \in S ~\&~ a,a' \in A.
    \end{split}
\end{equation}
The above gradient of $\log \pi_\theta$ is also provided in Equation 37 in Agarwal et al. 2019.
The above algorithm is a actor-critic type algorithm based on policy iteration which performs state-action value function update ($Q_{n}(s,a)$), variance function update ($\variance_{n}(s,a)$) and the policy parameter update ($\theta_{n}$).

\subsubsection{Sketch of convergence analysis}

Let \[\underset{k <n}{\forall} \F_n =  \sigma( \theta_k, Q_{k}(s,a),~\variance_{k}(s,a), ~Y_k(s,a),~\actionsample_{k}(Y_{k}(s,a)), ~a(k), ~b(k), ~c(k)) \] denote the associated sigma-field. Here $\sigma_k(.,.)$ denotes the state-action variance (notation overloading). Lets define now the sequence:

\begin{equation}\label{martingale_value}
\begin{split}
B^{(1)}_{s,a}(n) = \sum_{k=0}^{n-1} a(k)\Big[&\Big(r(s,a) + Q_{k}(Y_{k}(s,a), \actionsample_{k}(Y_{k}(s,a)))\Big) \\
& -\sum_{s' \in S, a' \in A} p(s'|s,a)\poltheta^{k}(a'|s')\Big(r(s,a) + Q_{k}(s',a')\Big) \Big].
\end{split}
\end{equation}
It is easily seen that $B^{(1)}_{s,a}(n)$ is martingale sequence with respect to filtration $\F_n, n \geq 0$. Let $M^{(1)}_{s,a}(n+1) = B^{(1)}_{s,a}(n+1) - B^{(1)}_{s,a}(n)$ denote the martingale difference sequence for $n \geq 0$. Similarly, lets define another sequence:
\begin{equation}\label{martingale_var}
\begin{split}
B^{(2)}_{s,a}(n) = \sum_{k=0}^{n-1} b(k)\Big[&\Big(\delta_{k}\big(s,a, Y_k(s,a),\actionsample_{k}(Y_{k}(s,a))\big)^2+ \variance_{k}\big(Y_k(s,a),\actionsample_{k}(Y_{k}(s,a))\big)\Big) \\
&-\sum_{s' \in S, a' \in A} p(s'|s,a)\poltheta^{k}(a'|s')\Big(\delta_k(s,a, s',a')^2+ \variance_{k}(s',a')\Big) \Big].
\end{split}
\end{equation}
such that $B^{(2)}_{s,a}(n)$ is martingale sequence with respect to filtration $\F_n, n \geq 0$. Let $M^{(2)}_{s,a}(n+1) = B^{(2)}_{s,a}(n+1) - B^{(2)}_{s,a}(n)$ denote the martingale difference sequence for $n \geq 0$. To see it is a martingale difference sequence, let \[\tau_k = \delta_{k}(s,a, Y_k(s,a),\actionsample_{k}(Y_{k}(s,a)))^2+ \variance_{k}(Y_k(s,a),\actionsample_{k}(Y_{k}(s,a))).\] Therefore, \[\E[\tau_k | \F_k] = \sum_{s' \in S, a' \in A} p(s'|s,a)\poltheta^{k}(a'|s')\Big(\delta_k(s,a,s',a')^2+ \variance_{k}(s',a')\Big).\] Expectation of martingale difference sequence, $\E\Big[M^{(2)}_{s,a}(n+1) \Big| \F_n\Big] = \E\Big[b(n)(\tau_n - \E[\tau_n \Big| \F_n])~| \F_n \Big] = 0$ satisfies Definition \ref{def:martingale_difference}. Let the another sequence for the policy be: 
\begin{equation}\label{martingale_policy}
\begin{split}
B^{(3)}_{s,a}(n) = \sum_{k=0}^{n-1} c(k)\Big[&\Big(\nabla_{\theta} \log \poltheta^{k}(a|s) (Q_{k}(s,a) - \psi \variance_{k}(s,a))\Big)\\ & -\sum_{s \in S}\statedist^{\poltheta^{k}}(s)\sum_{a \in A}\poltheta^{k}(a|s)\Big(\nabla_{\theta} \log \poltheta^{k}(a|s)(Q_{k}(s,a) - \psi \variance_{k}(s,a))\Big) \Big].
\end{split}
\end{equation}
which is a martingale sequence with respect to $\F_n, n\geq 0$. Here $\statedist^{\poltheta}(s) = \sum_{t=0}^{\infty} P(S_t = s  \mid s_0, \poltheta)$ is the stationary probability of a Markov Chain of being in a state $s \in S$ under $\poltheta$ policy. Let $M^{(3)}_{s,a}(n+1) = B^{(3)}_{s,a}(n+1) - B^{(3)}_{s,a}(n)$ denote the martingale difference sequence for $n \geq 0$ (can be shown as a martingale difference sequence easily similar to the one for $M^{(2)}_{s,a}(n+1)$ above).

Using \eqref{Q_stochastic} and \eqref{martingale_value}, the equation can be re-written as:
\begin{equation}\label{Q_full_stochastic}
\begin{split}
  Q_{n+1}(s,a) &=  Q_{n}(s,a) + a(n)\Big(\\&\sum_{s' \in S, a' \in A} p(s'|s,a)\poltheta^{n}(a'|s')\Big(r(s,a) + Q_{n}(s',a') - Q_{n}(s,a)\Big)\Big) + M^{(1)}_{s,a}(n+1).
\end{split}
\end{equation}
Similarly, using \eqref{variance_stochastic} and \eqref{martingale_var}, the equation for the variance is expressed as:
\begin{equation}\label{variance_full_stochastic}
\begin{split}
    \variance_{n+1}(s,a) &= \variance_{n}(s,a) + b(n) \Big( \sum_{s' \in S, a' \in A} p(s'|s,a)\poltheta^{n}(a'|s')\times\\&\quad \Big(\delta_n(s,a,s',a')^2+ \variance_{n}(s',a') - \variance_{n}(s,a)\Big)\Big) + M^{(2)}_{s,a}(n+1).
\end{split}
\end{equation}
Using, \eqref{policy_stochastic} and \eqref{martingale_policy}, the equation for the policy parameter is re-framed as:
\begin{equation}\label{policy_full_stochastic}
\begin{split}
    \theta_{n+1} &= \theta_{n} + c(n)\Big(\\&\sum_{s \in S, a \in A}\statedist^{\poltheta^{n}}(s)\poltheta^{n}(a|s)\Big(\nabla_{\theta} \log \poltheta^{n}(a|s)(Q_{n}(s,a) - \psi \variance_{n}(s,a))\Big)\Big) +  M^{(3)}_{s,a}(n+1).
    \end{split}
\end{equation}

 
To complete the proof of multi timescale analysis, here we will set up the notations to show the convergence of the online and tabular variance under a fixed policy $\pi$ and a fixed state-action value function $Q_{\pi}$ in Lemma \ref{lemma:online_TD_variance}. We would follow the notations of \cite{Bertsekas1996NP} and use concepts introduced in Section \ref{section:SA} and Proposition  \ref{prop:basic_sa_convergence} here.  The variance of return for a given policy $\pi$ is represented by: \[\sigma_\pi(s,a) = \E_{\pi}[\delta_t^2 + \sigma_\pi(s_{t+1}, a_{t+1}) \mid s_t = s,a_t= a]\] where $\delta_t$ is : 
\begin{align}\label{td_for_value}
  \delta_t(s_{t}, a_{t}, s_{t+1}, a_{t+1}) = r(s_t, s_{t+1}) + Q_\pi(s_{t+1}, a_{t+1}) - Q_\pi(s_{t},a_{t}).  
\end{align}
Here $Q_\pi$ is the state-action value function and $r$ is the reward contingent on  the current and the next state. We would be using $t$ for trajectory; $m \in \{0, 1, \dots, N_t\}$ for time step where $N_t$ is the maximum length of a trajectory; the stepsize $\gamma_{t}(i), ~i = \{1,2, \dots n\}$. 


The \textbf{online update} for the state-action value function ($\Q_{t,m}^{o}(s)$) using \tdzero method is given by:
\begin{equation}\label{variance_online_tdl_update}
    \begin{split}
        \Q_{t,0}^{o}(s,a) &= \Q_{t}^{o}(s,a) \quad \forall ~ s \in S, a \in A,\\
        \delta_{m,t}^{o} &= r(s_{m}^t, s_{m+1}^t) + \Q_{t,m}^{o}(s_{m+1}^t, a_{m+1}^t) - \Q_{t,m}^{o}(s_{m}^t, a_{m}^t),\\
         \Q_{t,m+1}^{o}(s,a) &= 
        \begin{cases}
        \Q_{t,m}^{o}(s,a) + \delta_{m,t}^{o},& \text{if } s = s_{m}, a = a_{m},\\
        \Q_{t,m}^{o}(s,a),              & \text{otherwise}.\\
        \end{cases}
        \\\Q_{t+1}^{o}(s,a) &= \Q_{t,N_t}^{o}(s,a) \quad \forall ~ s \in S, a \in A.
    \end{split}
\end{equation}

The \textbf{online update} for the variance ($\sigma_{t,m}^{o}(s)$) using \tdzero method is given by:
\begin{equation}\label{variance_online_tdl_update}
    \begin{split}
        \sigma_{t,0}^{o}(s,a) &= \sigma_{t}^{o}(s,a) \quad \forall ~ s \in S, a \in A,\\
        d_{m,t}^{o} &= \delta_{m,t}^2 + \sigma_{t,m}^{o}(s_{m+1}^t, a_{m+1}^t) - \sigma_{t,m}^{o}(s_{m}^t, a_{m}^t),\\
         \sigma_{t,m+1}^{o}(s,a) &= 
        \begin{cases}
        \sigma_{t,m}^{o}(s,a) + d_{m,t}^{o},& \text{if } s = s_{m}, a = a_{m},\\
        \sigma_{t,m}^{o}(s,a),              & \text{otherwise}.\\
        \end{cases}
        \\\sigma_{t+1}^{o}(s,a) &= \sigma_{t,N_t}^{o}(s,a) \quad \forall ~ s \in S, a \in A.
    \end{split}
\end{equation}
where $\delta_{m,t}^2$ is the square of equation \eqref{td_for_value}.

\begin{assumption}\label{assump:proper_policy}
The policy $\pi$ is proper (see Definition \ref{def:proper_policy}).
\end{assumption}


We will now establish how the updates \eqref{Q_full_stochastic} - \eqref{policy_full_stochastic} can be mapped to the three o.d.e. \eqref{x} -  \eqref{z} respectively. Here, we assume that we follow \eqref{step-size law} for the three step sizes we use in \eqref{Q_full_stochastic} - \eqref{policy_full_stochastic} along with the condition \eqref{stepsize_three_level}. As a direct consequence of \eqref{stepsize_three_level}, \eqref{Q_full_stochastic} moves at the fastest timescale, followed by \eqref{variance_full_stochastic}. \eqref{policy_full_stochastic} moves at the slowest timescale as compared to the two previous equations.
We will now satisfy the Assumptions \eqref{assump:MTA_1} - \eqref{assump:MTA_4} to claim that the above three-timescale actor-critic converges. 

\skippingparagraph\textbf{For requirement} \ref{assump:MTA_1}: The iterates of $\norm{Q_{n}}{}$,$\norm{\variance_{n}}{}$ already satisfy $\norm{Q_{n}}{} < \infty$ and $\norm{\variance_{n}}{} < \infty$. One can also easily see that $\norm{\theta_{n}}{} < \infty$ since \eqref{grad_log_pi} is bounded. Therefore, we have $\sup_{n}( \norm{Q_n}{} + \norm{\variance_n}{} + \norm{\theta_n}{}) < \infty$.

\skippingparagraph\textbf{For requirement} \ref{assump:MTA_2}: 
For a given fixed policy $\pi$, the proof of convergence of online and tabular critic (state-action value function $Q_n$) using \tdzero to a stationary point $Q_{\pi}$ with probability 1 is established using Assumption \eqref{assump:proper_policy}, \eqref{assump:reward_bounded}, \eqref{assump:TDerrorValue_bounded} and Lemma \ref{lemma:online_TD_value}.

\skippingparagraph\textbf{For requirement} \ref{assump:MTA_3}: For a given fixed policy $\pi$, the proof of convergence of a online, tabular variance of the return $\sigma_n$ to a stationary point $\sigma_\pi$ can be established by using Assumption \eqref{assump:proper_policy} and Lemma \ref{lemma:online_TD_variance} a.s..

\skippingparagraph\textbf{For requirement} \ref{assump:MTA_4}: The conditions can be relaxed to local rather than global stationary point because we want to just show the convergence of the policy. Under the optimal state-action value and variance functions for every step, it follows the policy-gradient and it converges to local fixed point.

 
\skippingparagraph This completes all the necessary requirements for the convergence of the proposed actor-critic algorithm using multi timescale analysis. The rest of the proof would follow through Lemma \ref{lemma:online_TD_variance} and Lemma \ref{lemma:online_TD_value}.

\begin{lemma}\label{lemma:online_TD_variance}
Considering the on-line \tdzero algorithm for the variance \eqref{variance_online_tdl_update} and the Assumptions \ref{assump:proper_policy} hold. Then $\sigma_{t,0}^{o}(s,a)$ converges to $\sigma^{\pi}(s,a)$ w.p. 1.
\end{lemma}

\begin{proof}
The stochastic approximation method for updating the offline variance is given as:
\begin{align}\label{var_sa}
    \sigma_{t+1}(s,a) = \sigma_{t}(s,a) + \gamma_t(s)\sum_{k \in W_{t}}d_{k,t},
\end{align}
where $W_t = \{m \mid s = s_{m}, a = a_{m}\}$ set of all the time steps in $t$ trajectory with same state and action pair $(s,a)$ and $d_{m,t}$ is the TD error for the variance using \eqref{td_for_value} is given by:
\begin{align}\label{td_for_offline_variance}
  d_{m,t} = \delta_{m,t}^2 + \sigma_t(s_{m+1}^t, a_{m+1}^t) - \sigma_t(s_{m}^t, a_{m}^t).  
\end{align}
Here $\delta_{m,t}^2$ is the pseudo-reward term like the usual one we have in the Bellman equation of the value function. To show the convergence analysis of online \tdzero for the variance function, we make the following assumptions.

\begin{assumption}
We have the true state-action value function for computing the TD error for the $Q$ values \[\delta_{m,t} = r(s_{m}^t, s_{m+1}^t) + Q_\pi(s_{m+1}^t, a_{m+1}^t) -Q_\pi(s_{m}^t, a_{m}^t).\]
\end{assumption}
This assumption is easy to satisfy, given we are following the three timescale architecture. The state-action value function iterates \eqref{Q_stochastic} is moving at a faster timescale compared to variance function iterates \eqref{variance_stochastic}. Due to faster convergence of the $Q$ iterates, for any given policy $\pi$, $\{Q_n \to Q(\pi)\}$ with probability 1. It follows the similar argument as the two timescale argument in \cite{borkar2009stochastic}. One can see $\{Q_n\}$ iterates as the one running in the inner loop (faster timescale) and $\{\sigma_n\}$ iterates in the outer loop (slower timescale). The inner loop sees the outer loop as quasi-static, while the latter sees the former as nearly equilibrated. Thus, when we are updating the iterates of the variance function, it is fair to assume that we have access to state-action value function $Q_\pi$ for some policy $\pi$.
\begin{assumption}\label{assump:reward_bounded}
The reward $|r(s_{m}^t, s_{m+1}^t)|$ is bounded by a constant P.
\end{assumption}

\begin{assumption}
$|\delta_{m,t}^2|$ is bounded above by a constant G.
\end{assumption}

\begin{assumption}\label{assump:TDerror_bounded}
The TD error for the variance function $|d_{m,t}|< (2 \norm{\sigma_t} + G)$ using \eqref{td_for_offline_variance}.
\end{assumption}

Similar to Sec 5.3 of \cite{Bertsekas1996NP}, we denote the offline update of variance by $\sigma_{t,m}(s)$ and the online update by $\sigma_{t,m}^{o}(s)$. The offline TD error contains $\sigma_{t}$ term, unlike $\sigma_{t,m}^{o}$ in an online update (see \eqref{td_for_offline_variance} and second line in \eqref{variance_online_tdl_update} for clarity) because the variance function is only updated after the completion of an entire trajectory in an offline update. The online update differs from offline update by a term which is of second order in the stepsize.

We would now bound the difference between the online and the offline update. Let $\hat \gamma_t = \max_s \gamma_t(s)$. We would assume that at the beginning of a trajectory both the online and the offline vectors have a same value i.e. $\sigma_{t}^{o}(s,a) = \sigma_{t}(s,a)$. Using the Assumption \ref{assump:TDerror_bounded},  we can write that the difference between the two terms of the offline method can be bounded as:
\begin{equation}\label{diff_offline_variance}
|\sigma_{t,m}(s,a) - \sigma_{t}(s,a)| \leq \gamma_t(s) N_t (2 \norm{\sigma_t} + G).    
\end{equation}
Similar to \cite{Bertsekas1996NP}, we would prove by induction that for some $D_m$ with $D_0= 0$:
\begin{align}\label{diff_online_offline_variance}
    |\sigma_{t,m}^{o}(s,a) -\sigma_{t,m}(s,a)| \leq D_m \hat\gamma_t \gamma_t(s)
\end{align}
The trivial case is when $m=0$ and $D_0 = 0$. Lets say for m the above statement holds. Now we will have to prove for case $m+1$. We know that,
\begin{align*}
    |\sigma_{t,m}^{o}(s,a) -\sigma_{t}(s,a)| &\leq |\sigma_{t,m}^{o}(s,a) -\sigma_{t,m}(s,a)| + |\sigma_{t,m}(s,a) -\sigma_{t}(s,a)| \\
    & = \{D_m \hat\gamma_t \gamma_t(s)\} + \{\gamma_t(s) N_t (2 \norm{\sigma_t} + G)\} \quad \text{Using \eqref{diff_online_offline_variance} \& \eqref{diff_offline_variance}}
\end{align*}
Therefore,
\begin{align}
    |d_{m,t}^{o} - d_{m,t}| &\leq 2 \norm{\sigma_{t,m}^{o} -\sigma_{t}}\\
    & = 2D_m \hat\gamma_t^2 + 2\hat\gamma_t N_t (2 \norm{\sigma_t} + G)
\end{align}

Now for $m+1$ case,
\begin{align*}
    |\sigma_{t,m+1}^{o}(s,a) -\sigma_{t,m+1}(s,a)| &\leq |\sigma_{t,m}^{o}(s,a) -\sigma_{t,m}(s,a)| + |\gamma_t(s) (d_{m,t}^{o} - d_{m,t}) |\\
    & \leq D_m \hat\gamma_t \gamma_t(s) + 2\hat\gamma_t \big(D_m \hat\gamma_t \gamma_t(s) + \gamma_t(s) N_t (2 \norm{\sigma_t} + G)\big)\\
    & \leq D_m \hat\gamma_t \gamma_t(s) (1 + 2\hat \gamma_t) + 2\hat\gamma_t \gamma_t(s) N_t (2 \norm{\sigma_t} + G)
\end{align*}

Therefore, the recursive form of $D_m$ can now be expressed as,
\begin{align*}
    D_{m+1} = D_m (1 + 2\hat \gamma_t) + 2 N_t (2 \norm{\sigma_t} + G).
\end{align*}
Using this, 
\begin{align*}
D_{N_t} &\leq N_t (1+2 \hat \gamma_t)^{N_t} 2 N_t (2 \norm{\sigma_t} + G) \\
& \leq A N_t^2 (1+2\hat \gamma_t)^{N_t}  (2 \norm{\sigma_t} + 1)
\end{align*}
for some constant A. Now the difference between the online and the offline update after the completion on a trajectory is expressed as,
\begin{align*}
    |\sigma_{t+1}^{o}(s,a) -\sigma_{t+1}(s,a)| & \leq |\sigma_{t,N_t}^{o}(s,a) -\sigma_{t,N_t}(s,a)| \leq D_{N_t}\hat\gamma_t \gamma_t(s)\\
    &\leq  A N_t^2 (1+2 \hat \gamma_t)^{N_t}  (2 \norm{\sigma_t} + 1)
\end{align*}

The update of the online variance algorithm is written as,
\[\sigma_{t+1}^{o}(s,a) = (1-\gamma_t(s))\sigma_{t}^{o}(s,a) + \gamma_t(s)\big((H_t \sigma_{t}^{o})(s,a) + w_t(s,a) + u_t(s,a)\big),\] where \[(H_t \sigma_{t}^{o}) (s,a) = \E[\sum_{k \in W_t} d_{k,t} \mid \F_t] + \sigma_{t}^{o} (s,a),\] and, \[w_t(s,a) = \sum_{k \in W_t} d_{k,t} - \E[\sum_{k \in W_t} d_{k,t} \mid \F_t].\] Here $\gamma_t(s) u_t(s,a) = \sigma_{t+1}^{o}(s,a) - \sigma_{t+1}(s,a)$. Therefore,

\begin{align*}
    |u_t(s)| \leq D_{N_{t}}\hat \gamma_t &\leq A \hat\gamma_t N_t^2 (1+2\hat \gamma_t)^{N_t}  (2 \norm{\sigma_t} + 1)
\end{align*}
Using Proposition \ref{prop:basic_sa_convergence}, in order to satisfy the condition on $u_t$, we need to show that $\theta_t = \hat\gamma_t N_t^2 (1+2 \hat \gamma_t)^{N_t}$ converges to $0$. Using Proposition 5.2 \cite{Bertsekas1996NP}, it can be easily shown that $\theta_t$ converge to 0. Using Lemma 5.2, Page 213 \cite{Bertsekas1996NP} we satisfy condition 2 of Proposition \ref{prop:basic_sa_convergence}. Further, using Lemma 5.3, Page 216 \cite{Bertsekas1996NP} we satisfy condition 3 of Proposition \ref{prop:basic_sa_convergence}. The proof is now complete, as we satisfy all the conditions of Proposition \ref{prop:basic_sa_convergence}, and we get that $\sigma_t$ converges to $\sigma_{\pi}$ w.p. 1.

\end{proof}

\begin{lemma}\label{lemma:online_TD_value}
Considering the on-line \tdzero algorithm for the state-action value function \eqref{variance_online_tdl_update} and the Assumptions \ref{assump:proper_policy} hold. Then $\Q_{t,0}^{o}(s,a)$ converges to $\Q^{\pi}(s,a)$ w.p. 1.
\end{lemma}
\begin{proof}
\begin{assumption}\label{assump:TDerrorValue_bounded}
The TD error for the Q function $|\delta_{m,t}|< (2 \norm{\Q_t} + P)$ using \eqref{td_for_value} and Assumption \ref{assump:reward_bounded}.
\end{assumption}

Using the similar line of arguments as of Lemma \ref{lemma:online_TD_variance}, the difference between the online and offline update of the state-action value function can be bounded using Assumption \ref{assump:TDerrorValue_bounded} as:

\begin{align*}
    |\Q_{t+1}^{o}(s,a) -\Q_{t+1}(s,a)| \leq B(1+2 \hat \gamma_t)^{N_t}  (2 \norm{\Q_t} + 1)
\end{align*}
for some constant $B$.

The update of the online state-action value function is written as,
\[\Q_{t+1}^{o}(s,a) = (1-\gamma_t(s))\Q_{t}^{o}(s,a) + \gamma_t(s)\big((H_t \Q_{t}^{o})(s,a) + w_t(s,a) + u_t(s,a)\big),\] where \[(H_t \Q_{t}^{o}) (s,a) = \E[\sum_{k \in W_t} \delta_{k,t} \mid \F_t] + \Q_{t}^{o} (s,a),\] and, \[w_t(s,a) = \sum_{k \in W_t} \delta_{k,t} - \E[\sum_{k \in W_t} \delta_{k,t} \mid \F_t].\] Here $\gamma_t(s) u_t(s,a) = \Q_{t+1}^{o}(s,a) - \Q_{t+1}(s,a)$. Therefore, using similar arguments as Lemma \ref{lemma:online_TD_variance}, we satisfy all the conditions of Proposition \ref{prop:basic_sa_convergence}. This completes the proof, we get $Q_t$ converges to $Q_{\pi}$ w.p. 1.
\end{proof}

\section{Pseudo algorithms}\label{app:pseudoalgo}

\subsection{Off-policy variance penalized actor-critic algorithm}
The pseudo-algorithm \ref{alg:off_VPAC} shows the prototype implementation of off-policy \texttt{VPAC}.
{
	\begin{algorithm}[h]
		\caption{Off-policy VPAC}
		\label{alg:off_VPAC}
		\begin{algorithmic}[1]
			\STATE Here $\alpha_w, \alpha_\theta, \alpha_z$ stands for step size of critic, policy and variance respectively. $b, \pi$ is behavioral and target policy respectively.
			\STATE \textbf{Input}: differentiable policy $\pi_{\boldsymbol{\theta}}(a|s)$, value $\hat{Q}(s,a,\boldsymbol{w})$, and variance $\hat{\sigma}(s,a,\boldsymbol{z})$
			\STATE \textbf{Parameters}: $\gamma \in [0,1]$, $\psi \in (0, \infty)$, $\alpha_\theta \textless \alpha_z  \textless \alpha_w $, $\bar{\gamma}=\gamma^2$
			\newline
			\STATE Initialize the parameters: $\theta, w, z$
			\FOR{Episode $i= 1,2,3 \dots $}
    			\STATE Initialize state $S$. Sample $A \sim b(.|S)$.
    			\STATE $I_Q, I_{\sigma} = 1, 1$
    			\STATE $\rho_Q, \rho_{\sigma} = \rho(S,A), \rho(S,A)$
    			\STATE $\bar{\psi} = \psi$
    			\STATE $\text{firstTimeStep} = True$  (flag variable)
    			\REPEAT
    			\STATE Observe $\{R, S'\}$. $A' \sim b(.|S')$
    			\STATE $\delta \gets \, R + \gamma\rho(S',A')\hat{Q}(S',A',w) - \hat{Q}(S,A, w)$
    			\STATE $\bar{\delta} \gets \, \delta^2 + \bar{\gamma}\rho(S',A')^2\hat{\sigma}(S',A',z) - \hat{\sigma}(S,A, z)$
    			\STATE $w \gets w +\alpha_w \delta \nabla_w \hat{Q}(S,A,w)$
    			\STATE $z \gets z +\alpha_z \bar{\delta} \nabla_z \hat{\sigma}(S,A,z)$
    			\STATE $\theta \gets \theta +\alpha_\theta \nabla_\theta \log(\pi_\theta(A|S)) \Big(I_Q \rho_Q \hat{Q}(S,A,w) - \bar{\psi} I_{\sigma}\rho_\sigma \hat{\sigma}(S,A,z)\Big)$
                \STATE $S \gets S', A \gets A'$
                \STATE if $(\text{firstTimeStep}==True):\bar{\psi} = 2\psi $, $\text{firstTimeStep} = False$
                \STATE $I_{Q} = \gamma I_{Q}$
    			\STATE $I_{\sigma} = \bar\gamma I_{\sigma}$
    			\STATE $\rho_Q \gets \rho(S,A)\rho_Q$
    			\STATE $\rho_\sigma \gets \rho(S,A)^2\rho_\sigma$
                \UNTIL{$S'$ is a terminal state}
            \ENDFOR
		\end{algorithmic}
	\end{algorithm}
}

\subsection{Baselines - Variance adjusted actor-critic (VAAC)}
Here, we are implementing \cite{tamar2013variance} as a baseline to compare with our proposed method. We are writing the notations with discounted setting (un-discounted originally in paper) for the clarity. $M(s) = \E[G_t^2|S_0 = s]$ is the second moment of the return and $Var(S) = \text{Var}[G_t|s_0 = s]$ is denoted as the variance in the return.
\begin{equation*}
\begin{split}
    M(s) &= \E[G_t^2 | S_0 = s]\\
    &= \E[(\sum_{t=0}^{T} \gamma^t r(S_t))^2 \mid S_0 = s]\\ 
    &=\E[(r(S_0) + \sum_{t=1}^{T} \gamma^t r(S_t))^2 \mid S_0 = s]\\
    &=r(s)^2 + \gamma^2 \sum_{s'}P(s'|s)M(s') + 2\gamma r(s) \sum_{s'}P(s'|s)V(s')
\end{split}
\end{equation*}

The objective function where $\mu$ is the variance regularizer: \[\eta(\theta)= V(S_0) - \mu Var(S_0), \text{ where } Var(S_0) = M(S_0) - V(S_0)^2 \text{ is the variance}.\]
\[\grad Var(S_0) = \grad M(S_0) - 2 V(S_0) \grad V(S_0).\]
Gradient of $M(s,a)$ is,
\begin{equation*}
    \begin{split}
        M(s,a) &= r(s,a)^2 + \gamma^2 \sum_{s'}P(s'|s,a)M(s') + 2\gamma r(s,a) \sum_{s'}P(s'|s,a)V(s')\\
    \grad M(s,a) &= \gamma^2 \sum_{s'}P(s'|s,a)\grad M(s') + 2\gamma r(s,a) \sum_{s'}P(s'|s,a)\grad V(s')\\
    \sum_{a} \policy(a|s) \grad M(s,a) &= \gamma^2 \sum_{s'}P(s'|s)\grad M(s') + 2\gamma \sum_{a} \policy(a|s) r(s,a) \sum_{s'}P(s'|s,a)\grad V(s').
    \end{split}
\end{equation*}

Gradient of $M(s)$ where $\bar q(s,a)$ is $\gamma^2$-discounted state-action visitation under policy $\policy$ is,
\begin{equation*}
    \begin{split}
        \grad M(s) &= \sum_{a}\grad \policy(a|s) M(s,a) + \sum_{a}\policy(a|s) \grad M(s,a)\\
        &=\sum_{a}\grad \policy(a|s) M(s,a) + \gamma^2\sum_{a}\policy(a|s) \sum_{s'}P(s'|s)\grad M(s') \\&\quad + 2\gamma \sum_{a} \policy(a|s) r(s,a) \sum_{s'}P(s'|s,a)\grad V(s')\\
        & = \sum_{s,a}\bar q(s,a) \grad \log \policy(s,a) M(s,a) + 2 \gamma \sum_{s,a, s'}\bar q(s,a) r(s,a) P(s'|s,a) \grad V(s').
    \end{split}
\end{equation*}
We now present the pseudo-algorithm \ref{alg:VAAC} for \texttt{VAAC} proposed by \cite{tamar2013variance}.

{
	\begin{algorithm}
		\caption{VAAC}
		\label{alg:VAAC}
		\begin{algorithmic}[1]
			\STATE $\alpha_w, \alpha_z, \alpha_\theta$ step size of value critic, variance critic and policy respectively.
			\STATE $Q(s,a) = w_{Q}^T \phi_{Q}(s, a)$ and $M(s,a) = w_{M}^T \phi_{M}(s, a)$.
			\STATE \textbf{Input}: a differentiable policy $\pi_{\boldsymbol{\theta}}(a|s)$
			\STATE \textbf{Parameters}:$\gamma \in [0,1]$, $\mu \in (0, \infty)$, step sizes -- $\alpha_w, \alpha_z, \alpha_\theta$.
			\STATE Initialize parameters: $\theta, w_Q, w_M, V(S_0)$
			\newline
			\FOR{Episode $i = 1,2, \dots$}
                \STATE Generate an episode $S_{0}, A_{0}, R_{1}, \dots, S_{T-1}, A_{T-1}, R_{T}$ following $\pi_{\theta_i}$.
                \STATE $w_{Q, i} = w_{Q}; w_{M, i} = w_{M}$.
                \STATE $\bar G = [0]$.
                \FOR{Step $t = 0,1, \dots, T-1 $}
                    \STATE $G = \sum_{k=t+1}^{T} \gamma^{k-(t+1)} R_k$.
                    \STATE $\bar G.\texttt{apppend}(\sum_{k=t+1}^{T} \gamma^{2(k-(t+1))} R_k)$.
                    \STATE $w_Q \gets w_Q + \alpha_z \Big[ G - w_{Q,i}^T \phi_{Q}(S_t, A_t)\Big]\phi_{Q}(S_t, A_t)$
                    \STATE $w_M \gets w_M + \alpha_w\Big[ G^2 - w_{M,i}^T \phi_{M}(S_t, A_t)\Big]\phi_{M}(S_t, A_t)$
                    \IF{$t==0$}
                        \STATE $V(S_0) \gets  V(S_0) + \alpha \Big[ G - V(S_0)\Big]$
                    \ENDIF
                \ENDFOR
                \STATE $\theta \gets \theta + \alpha_\theta \Bigg[\sum_{t=0}^{T} \grad_\theta \log \pi(A_t| S_t)\Bigg\{ \gamma^t Q(S_t, A_t) - \mu \Big(\gamma^{2t} M(S_t, A_t) + 2 \gamma^{t+1} \bar G[t]Q(S_t, A_t) - 2 \gamma^{t} V(S_0) Q(S_t, A_t) \Big) \Bigg\}\Bigg]$
            \ENDFOR
		\end{algorithmic}
	\end{algorithm}
}

For fair comparison with our proposed algorithm \texttt{VPAC} (with TD critic for value and variance function), we also converted \texttt{VAAC} algorithm \cite{tamar2013variance} into a TD style update for both $Q$ and $M$ function. We present Algorithm \ref{alg:VAAC_TD} as \texttt{VAAC\_TD} with TD updates for the critics.
{
	\begin{algorithm}[]
		\caption{VAAC\_TD}
		\label{alg:VAAC_TD}
		\begin{algorithmic}[1]
			\STATE Here $\alpha_w, \alpha_z, \alpha_\theta$ stands for step size of value critic, variance critic, and policy respectively.
			\STATE \textbf{Input}: a differentiable policy $\pi_{\boldsymbol{\theta}}(a|s)$, value $\hat{Q}(s,a,\boldsymbol{w})$, second order return $\hat{M}(s,a,\boldsymbol{z})$
			\STATE \textbf{Parameters}:$\gamma \in [0,1]$, $\mu \in (0, \infty)$,  $\alpha_\theta \textless \textless \alpha_w, \alpha_z$
			\STATE Initialize the parameters: $\theta, w, z$
			\newline
			
			\FOR{Episode $i = 1, 2,3, \dots$}
    			\STATE Initialize state $S$. Sample $(A\sim \pi_\theta(.|S))$.
    			\STATE $S_0 = S$ (record initial state)
    			\STATE $I_Q, I_{M} = 1, 1$
    			\STATE $G = 0$
    			\REPEAT
    			\STATE Observe $\{R, S'\}$. $A' \sim \pi_\theta(.|S')$
    			\STATE $\delta \gets \, R + \gamma\hat{Q}(S',A',w) - \hat{Q}(S,A, w)$
    			\STATE $\bar{\delta} \gets \, R^2 + \gamma^2 \hat{M}(S',A',z) + 2\gamma R  \hat{Q}(S,A, w)$
    			\STATE $w \gets w +\alpha_w \delta ~\nabla_w \hat{Q}(S,A,w)$
    			\STATE $z \gets z +\alpha_z \bar{\delta} ~\nabla_z \hat{M}(S,A,z)$
    			\STATE $V_0 = \sum_a \pi_\theta(a|S_0) \hat{Q}(S_0,a,w)$
    			\STATE $\theta \gets \theta +\alpha_\theta \nabla_\theta \log(\pi_\theta(A|S))\Big(I_{Q}\hat{Q}(S,A,w)- \mu \big\{I_M \hat{M}(S,A,z) + (2\gamma I_Q G \times \hat{Q}(S,A,w)) - (2 I_Q V_0 \times\hat{Q}(S,A,w)) \big\}\Big)$
    			\STATE $I_{Q}, I_{M} = \gamma I_{Q}, \gamma^2 I_{M}$
    			\STATE $G = G + I_{M} R$
    			\STATE $S \gets S', A \gets A'$
    			\UNTIL{$S'$ is a terminal state}
			\ENDFOR
		\end{algorithmic}
	\end{algorithm}
}

\section{Experiments}\label{app:experiment}
The code for all the experiments are added in the \textbf{Github}\footnote{ Code for the experiments are available \url{https://github.com/arushi12130/VariancePenalizedActorCritic.git}.}. We present here all the hyperparameters used for all the algorithms in the main paper Experiment section.

\subsection{On-policy VPAC}
\begin{figure}[h]
		\begin{center} \includegraphics[width=0.7\textwidth]{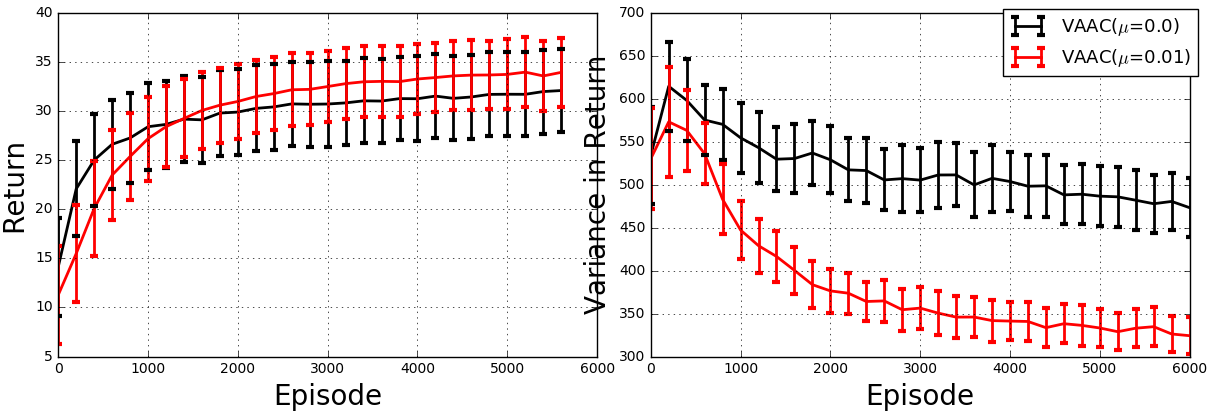}
        \caption[]
        {\textbf{Baseline \texttt{VAAC} performance in fourrooms environment}: Shows the performance of baseline \texttt{VAAC} \cite{tamar2013variance} in terms of mean and variance in the return in tabular fourrooms environment, where $\mu$ is the mean-variance tradeoff parameter. As $\mu$ increases, the variance decreases (see the right plot).}
        \label{fig:fr_vaac_performance}
        \end{center}
\end{figure}

\begin{figure}[!h]
        \centering
		\begin{subfigure}[b]{0.45\linewidth}
			\centering
			\captionsetup{justification=centering}
			\includegraphics[width=0.8\textwidth]{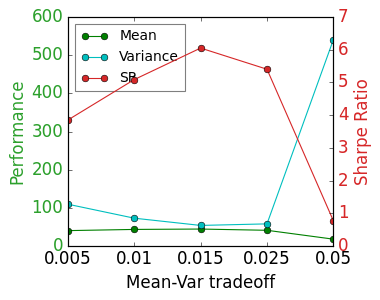}
			\caption[]{\small{VPAC}}
		\end{subfigure}
		\begin{subfigure}[b]{0.45\linewidth}
			\centering
			\captionsetup{justification=centering}
			\includegraphics[width=0.8\textwidth]{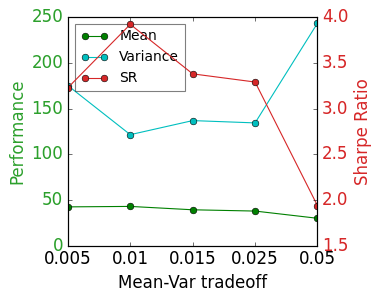}
			\caption[]{\small{VAAC\_TD}}
		\end{subfigure}
		\caption[]{\textbf{Effect of varying mean-variance tradeoff parameter on performance in four rooms}: Show the mean (green) and the variance (cyan) performance averaged over $50$ runs for both \texttt{VPAC} and \texttt{VAAC\_TD} by varying the mean-variance tradeoff $\psi$ and keeping all other hyperparameters (step sizes, discount factor, temperature) fixed. The red line shows the Sharpe Ratio $\frac{\E[G]}{\sqrt{Var[G]}}$ to view the combine effect of mean and variance.} 
        \label{fig:psi_performance}
\end{figure}

\begin{figure}[!h]
        \centering
		\begin{subfigure}[b]{0.45\linewidth}
			\centering
			\captionsetup{justification=centering}
			\includegraphics[width=0.75\textwidth]{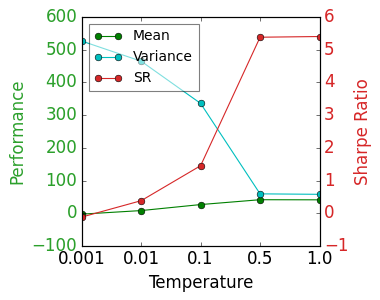}
			\caption[]{\small{VPAC}}
		\end{subfigure}
		\begin{subfigure}[b]{0.45\linewidth}
			\centering
			\captionsetup{justification=centering}
			\includegraphics[width=0.75\textwidth]{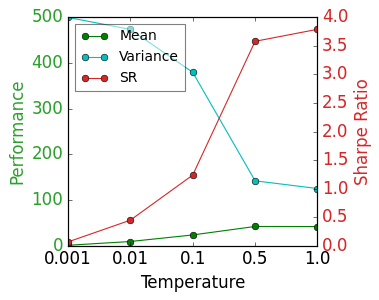}
			\caption[]{\small{VAAC\_TD}}
		\end{subfigure}
		\caption[]{\textbf{Effect of varying hyperparameter temperature on performance in four rooms}:  Show the mean (green) and the variance (cyan) performance averaged over $50$ runs for both \texttt{VPAC} and \texttt{VAAC\_TD} by varying the temperature (temp) parameter and keeping all other hyperparameters (step sizes, discount factor, $\psi$) fixed. The red line shows the Sharpe Ratio $\frac{\E[G]}{\sqrt{Var[G]}}$ to view the combine effect of mean and variance.} 
        \label{fig:temp_performance}
\end{figure}

\begin{figure}[]
        \centering
		\begin{subfigure}[b]{0.8\linewidth}
			\centering
			\captionsetup{justification=centering}
			\includegraphics[width=0.95\textwidth]{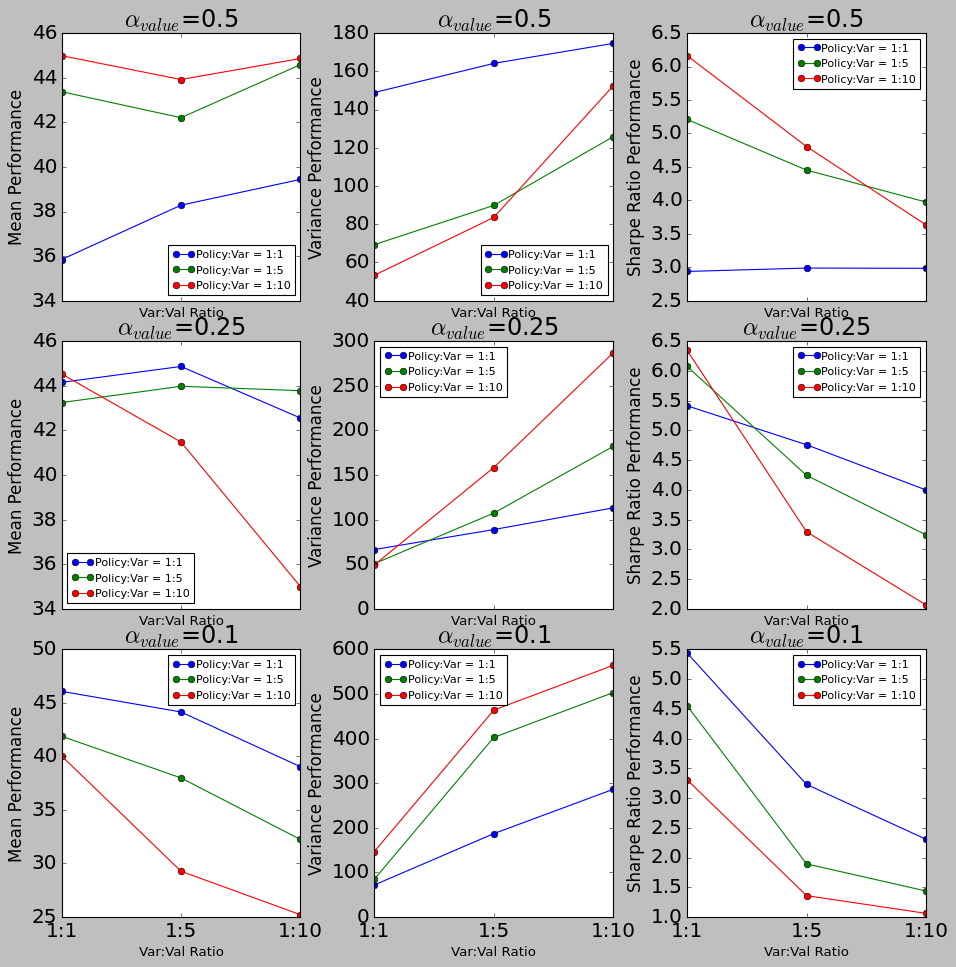}
			\caption[]{\small{VPAC}}
		\end{subfigure}
		\hspace{\fill}
		\hspace{\fill}%
		\begin{subfigure}[b]{0.8\linewidth}
			\centering
			\captionsetup{justification=centering}
			\includegraphics[width=0.95\textwidth]{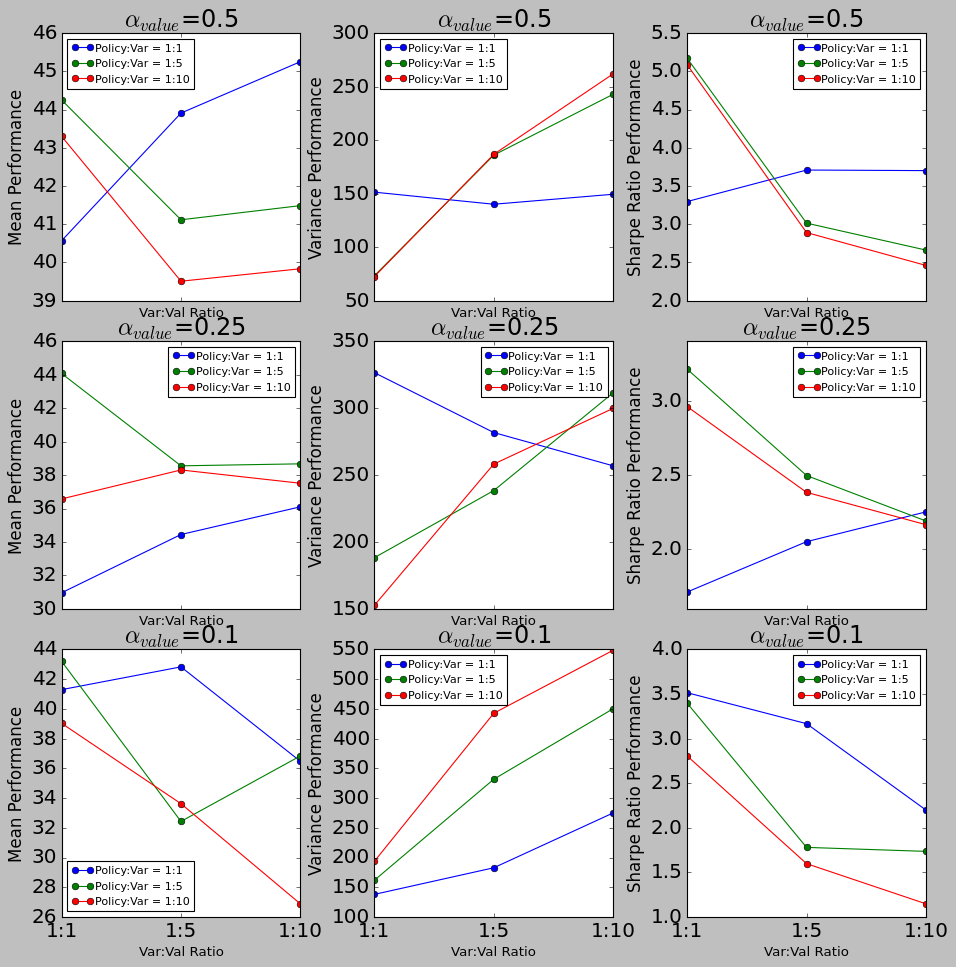}
			\caption[]{\small{VAAC\_TD}}
		\end{subfigure}
		\caption[]{\textbf{Step size sensitivity analysis}: The performance - mean, variance, Sharpe ratio (three columns respectively)- averaged over $50$ runs as a function of varying the ratio variance:value step size on x-axis and policy:variance step size shown in solid lines in the legend. We show the performance analysis for the three different value step sizes $\{0.5, 0.25, 0.1\}$ as three rows in the plot. The analysis is shown for (a) \texttt{VPAC} and (b) \texttt{VAAC\_TD} by keeping other hyperparameters (discount, temperature) fixed.}
		\label{fig:step_size_sensitivity}
\end{figure}

\paragraph{Tabular environment}
Table \ref{tab:discrete_on_policy} shows the hyper-parameters used for the results reported in the main paper. In the Fig. \ref{fig:fr_vaac_performance}, the mean-variance performance of the baseline \texttt{VAAC} \cite{tamar2013variance} is shown, where, $\mu=0$ depicts the performance without variance tradeoff (black line) and $\mu>0$ shows the performance on applying the penalization on the variance in return (red line). The Fig. \ref{fig:psi_performance} shows the effect of varying mean-variance tradeoff on the mean and the variance performance for \texttt{VPAC} and \texttt{VAAC\_TD}. We also show the Sharpe Ratio \cite{sharpe1994sharpe} for ease. We keep all other hyperparameters fixed. Similarly, in the Fig. \ref{fig:temp_performance} we show the effect of varying temperature and keeping other hyperparameters fixed on the performance. In Fig. \ref{fig:step_size_sensitivity} we present the step size sensitivity analysis for \texttt{VPAC} and \texttt{VAAC\_TD} on the performance metric - mean, variance and Sharpe Ratio. Since, the proposed algorithm is based on three-timescales namely -- value $\alpha_{w}$, variance $\alpha_{z}$ and policy $\alpha_{\theta}$, we present the effect of varying the ratios $\frac{\alpha_z}{\alpha_w}$ and $\frac{\alpha_\theta}{\alpha_z}$ on the performance. In the figure, we present the analysis with three different step size of value $\alpha_w = \{0.5, 0.25, 0.1\}$ as the three rows. On the x-axis, we vary $\frac{\alpha_z}{\alpha_w}$ (variance:value) as $\{1:1, 1:5, 1:10\}$, where the step size of variance is smaller or equal to that of value function. The solid lines in the plot (legend) show the performance with $\frac{\alpha_\theta}{\alpha_z}$ (policy:variance) as $\{1:1, 1:5, 1:10\}$, where step size of policy is smaller or equal to variance function. When the step size of value is higher (see the top row), we observe better performance  with same step size of variance but a smaller step size of policy (see performance of 1:10 policy:variance step size). As $\alpha_{w}$ decreases (bottom row), we observe that higher policy step size (see 1:1 policy:variance step size) performs better because growth rate diminishes with a very small policy step size. We observe similar behavior for both \texttt{VPAC} and \texttt{VAAC\_TD}.

\begin{table}[h]
\caption{Tabular four rooms experiments with on-policy VPAC and baseline VAAC, VAAC\_TD}
\label{tab:discrete_on_policy}
\centering
\begin{tabular}{ccccccc}
\hline
\textbf{Algorithm} & \textbf{$\psi$} & \textbf{$\alpha_\theta$ (policy)} & \textbf{$\alpha_w$ (value)} & \textbf{$\alpha_z$ (variance)} & \textbf{$\gamma$}  & \textbf{temp} \\\hline
\textit{AC}      & 0.0          & 0.01           & 0.5             & -                 & 0.99                          & 1             \\
\textit{VPAC}      & 0.015        & 0.01           & 0.5             & 0.5               & 0.99                         & 1             \\
\textit{VAAC\_TD}  & 0.01         & 0.01           & 0.5             & 0.5               & 0.99                         & 1             \\
\textit{VAAC}      & 0.01         & 0.001          & 0.05            & 0.005             & 0.99                         & 1            \\
\hline
\end{tabular}
\end{table}


\begin{figure}[]
        \begin{center}
        \includegraphics[width=0.9\linewidth]{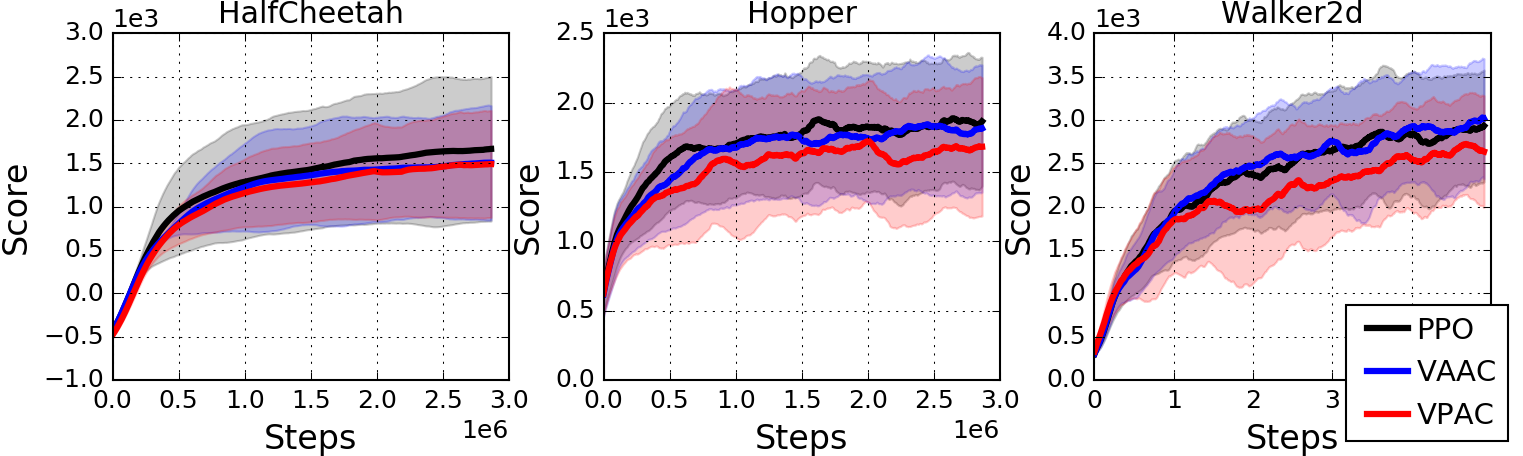}
        \caption[]
        {\textbf{Learning curves in Mujoco}: Compares the mean performance averaged over multiple random seeds for different environments in \texttt{PPO,VAAC} and \texttt{VPAC}. The $\psi$ value of \texttt{VPAC} algorithm are $0.15, 0.2, 0.15$ on HalfCheetah, Hopper and Walker2d respectively. \texttt{VAAC} uses $\psi$ value $0.15$ in all the three environments.} 
        \label{fig:mujoco_learningcurve}
        \end{center}
\end{figure}

\paragraph{Continuous state-action domain (MuJoCo)}
To incorporate the variance penalization objective function, a separate network for the variance is added in the \texttt{PPO} framework. We represent the approximation for variance by $\sigma(s,\boldsymbol{z})$ for $\forall s \in \mathcal{S}$ which is parameterized by $z$. A separate network similar to the value function was built to estimate the variance of a state using the same hyper-parameter settings as the value model. The code for the experiments is available on Github. The hyperparameters used in these experiments are mentioned in the code repository. The target estimate of the variance is given by $\hat{R}_{\sigma(t)}= \sum_{l=0}^{\infty} \bar{\gamma}^l\delta_{t+l}^2$, where $\delta_t$ is the one-step TD error at time $t$. We used a squared loss function for improving the estimates. We use a generalized advantage estimator (GAE) \citep{schulman2015high} to estimate an exponentially weighted average of the $k$-step advantage estimator for both the value and the variance functions. Let the one-step error in the variance be $\bar{\delta}_t^\sigma = \delta_t^2 + \bar{\gamma} \sigma(s_{t+1},z) - \sigma(s_t,z)$. Following the derivations of \cite{schulman2015high}, the GAE for variance is $A_{\sigma}^{GAE(\bar{\gamma}, \lambda)} = \sum_{l=0}^{\infty} (\bar{\gamma}\lambda)^l \bar{\delta}_{t+l}^\sigma.$
The variance-penalized objective function following \cite{schulman2017proximal}, where $A_{V}^{GAE(\gamma, \lambda)}$ represents GAE for the value is defined as follows:
\begin{equation*}
    \begin{split}
        J_{\text{VPAC}}(\theta) =& \min\Bigg(\frac{ \pi_{\theta}(A|S)}{\pi_{\text{old}}(A| S)} A_{\text{VPAC}},  \text{clip}\Big(\frac{ \pi_{\theta}(A|S)}{\pi_{\text{old}}(A| S)}, 1-\epsilon, 1+\epsilon \Big)A_{\text{VPAC}} \Bigg),\\
        &\text{where, }  A_{\text{VPAC}} = \Big(A_{V}^{\text{GAE}(\gamma, \lambda)} \! - \psi  A_{\sigma}^{\text{GAE}(\bar{\gamma}, \lambda)} \Big).
    \end{split}
\end{equation*}
Here $\epsilon$ is a hyperparameter, which decides how far a new policy can go away from the old policy. Fig. \ref{fig:mujoco_learningcurve} shows the mean performance along the learning curve for the three algorithms \texttt{VPAC, PPO, VAAC\_TD}, where the shaded region show the distribution of mean return across runs. $\psi$ is the only relevant hyperparameter in our method and we've explored it's effect in our experiments by keeping all the other hyper-parameters same (for all algorithms). The optimal $\psi$ varies based on benchmarks \& algorithms.

\subsection{Off-policy VPAC}

\begin{figure}[th]
		\begin{center}
        \begin{subfigure}[b]{0.30\textwidth}
            \centering
            \captionsetup{justification=centering}            \includegraphics[width=0.7\linewidth]{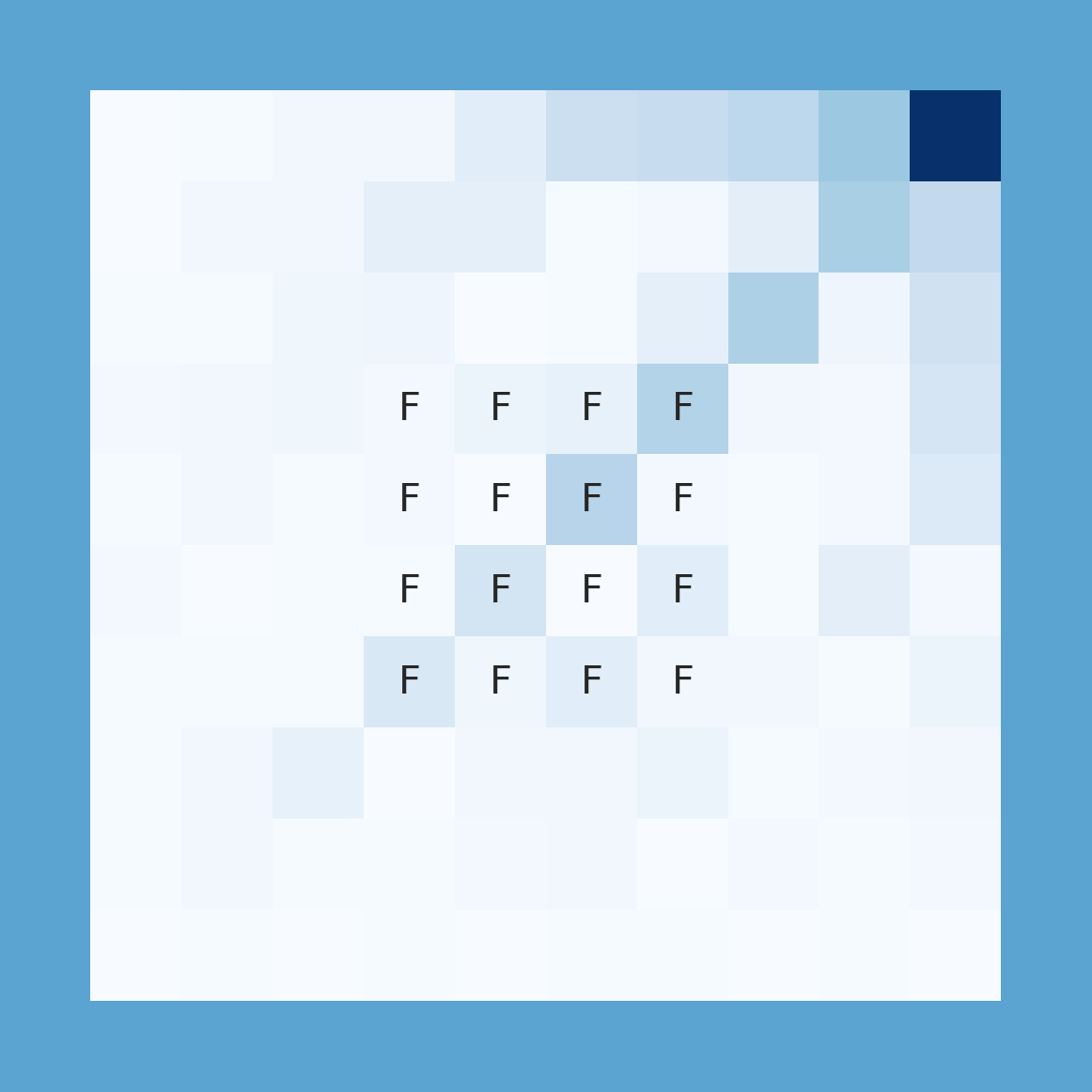}
            \caption[]%
            {{\small AC}} 
        \end{subfigure}
        \quad
        \begin{subfigure}[b]{0.30\textwidth}  
            \centering 
            \captionsetup{justification=centering}            \includegraphics[width=0.7\linewidth]{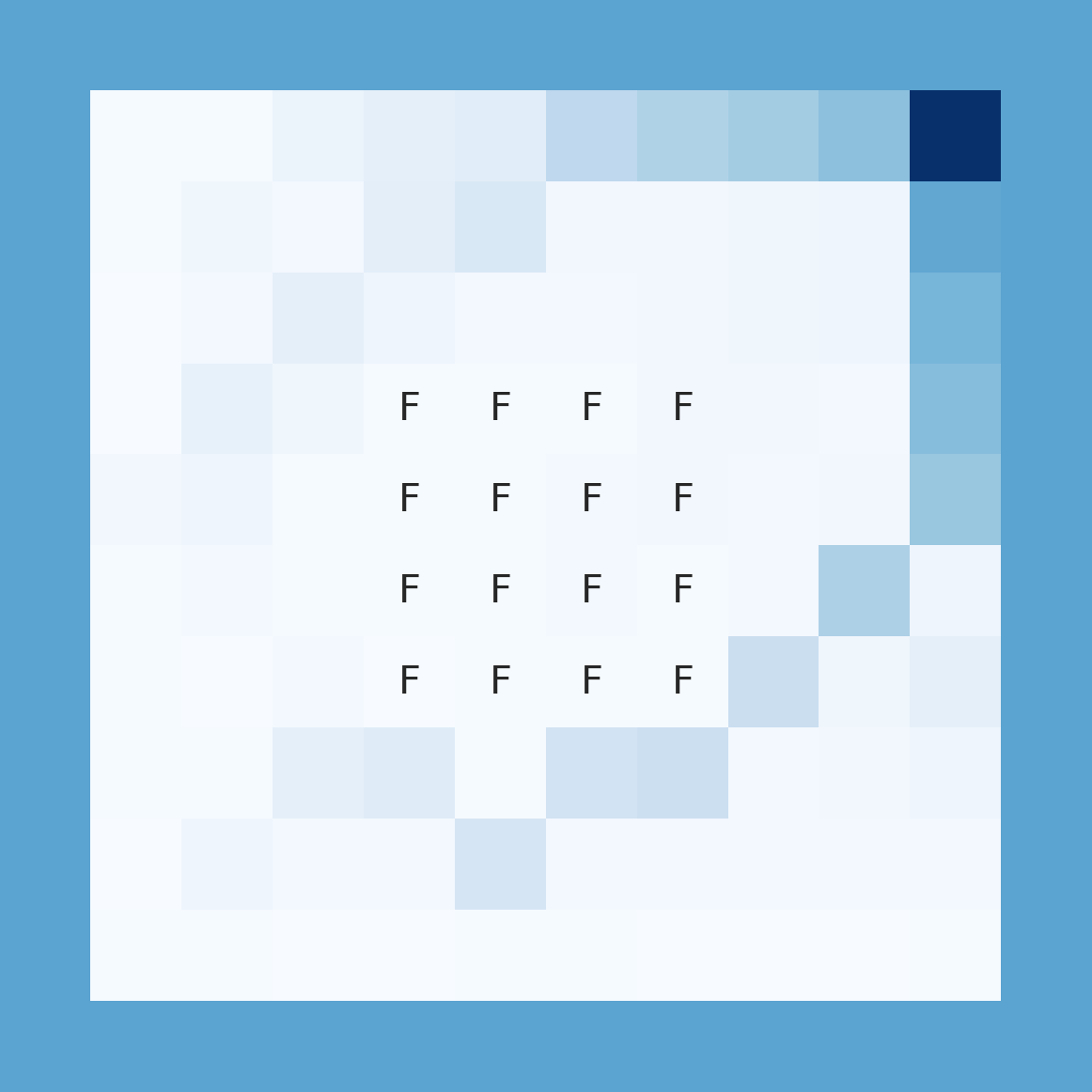}
            \caption[]%
            {{\small VPAC}} 
        \end{subfigure}
        \caption[]
        {\textbf{State visitation frequency in discrete puddle-world environment}: Frequency of state visits in both off-policy (a) AC, and (b) VPAC. Darker shade represents higher values. The states with $F$ depicts the variable reward receiving puddle region.} 
        \label{fig:freq_puddle_discrete}
        \end{center}
\end{figure}

\paragraph{Tabular environment}
Table \ref{tab:discrete_off_policy} shows the hyper-parameters of the results reported for the  tabular off-policy experiments. Agent in the discrete puddle-world environment, receives a reward from $\mathcal{N}(\mu=0, \sigma=8)$ distribution for the puddle region placed in the center. Normal (non-puddle region) receives a $0$ reward, whereas, on reaching the goal state (rightmost corner) gets a $50$ reward. Fig. \ref{fig:freq_puddle_discrete} compares the frequency of the state visitation with both the baseline \texttt{AC} and our proposed method \texttt{VPAC}. Darker shade represents higher value. \texttt{VPAC} learns to avoid the variable reward puddle-region denoted by $F$ in the figure, whereas, $AC$ takes the shortest path in order to reach the goal state.

\begin{table}[h]
\caption{Tabular puddle-world experiment with off-policy VPAC}
\label{tab:discrete_off_policy}
\centering
\begin{tabular}{ccccccc}
\hline
\textbf{Algorithm} & \textbf{$\psi$} & \textbf{$\alpha_\theta$ (policy)} & \textbf{$\alpha_w$ (value)} & \textbf{$\alpha_z$ (variance)} & \textbf{$\gamma$} &  \textbf{temp} \\\hline
\textit{AC}      & 0.0          & 0.05           & 0.5             & -                 & 0.99                          & 100             \\
\textit{VPAC}      & 0.002        & 0.05           & 0.5             & 0.25               & 0.99                          & 100            \\
\textit{VAAC}      & 0.0015        & 0.005           & 0.01             & 0.01               & 0.99                          & 100            \\
\hline
\end{tabular}
\end{table}

\paragraph{Continuous state domain} 
We use the similar tabular puddle-world environment, apart from changing it to continuous 2-dimensional environment in $[0,1]^2$. The action space consists of $4$ actions which change the position by $0.05$ in any one of the coordinates. In each transition, noise drawn from $\textit{Uniform}[-0.025, 0.025]$ distribution is added in both the coordinates. We added a variable reward puddle region with a centre at $(0.5, 0.5)$ and $0.4$ height with reward distribution as $\mathcal{N}(\mu=0, \sigma=8)$. An agent can be initialized from any state and the goal is placed at $(1,1)$. Whenever an agent is within a L$1$ norm of $0.1$ from the current state to the goal state, agent receives a reward of $50$. The agent receives the reward of $0$ from the normal states (non-puddle region). On expectation, the reward received for both the normal and the puddle regions is the same. Due to this, the agent learning with conventional objective of maximizing the mean performance has no benefit of avoiding the variable reward puddle region. The behavior policy is uniform over all the actions. The target policy is a Boltzmann distribution over the policy parameters. We use Retrace for the correction factor. The agent can take a maximum of $5000$ steps. The  standard tile coding \citep{sutton1998reinforcement} is used for discretization of the state-space. We used $10$ tilings, each of $5 \times 5$ over the joint space of $2$ features which is hashed to a $1024$ dimension vector. The hyperparameters for the reported results in the paper are presented in Table \ref{tab:cont_puddle}.
\begin{table}[th]
\caption{Continuous puddle-world experiment with off-policy VPAC}
\label{tab:cont_puddle}
\centering
\begin{tabular}{ccccccc}
\hline
\textbf{Algorithm} & \textbf{$\psi$} & \textbf{$\alpha_\theta$ (policy)} & \textbf{$\alpha_w$ (value)} & \textbf{$\alpha_z$ (variance)} & \textbf{$\gamma$}  & \textbf{temp} \\\hline
\textit{AC}      & 0.0          & 0.1           & 0.5             & -                 & 0.99                          & 50             \\
\textit{VPAC}      & 0.001        & 0.1           & 0.5             & 0.25               & 0.99                         & 50           \\
\textit{VPAC}      & 0.005        & 0.1           & 0.5             & 0.25               & 0.99                         & 50           \\
\textit{VAAC}      & 0.0015        & 0.005           & 0.05             & 0.05               & 0.99                          & 50           \\
\hline
\end{tabular}
\end{table}

\FloatBarrier